%% file: CameraReady2027.tex
\documentclass[letterpaper]{article}
\usepackage{aaai2027}

\usepackage[hyphens]{url}
\usepackage{graphicx}
\usepackage{natbib}
\usepackage{caption}
\usepackage{algorithm}
\usepackage{algorithmic}
\usepackage{newfloat}
\usepackage{listings}
\DeclareCaptionStyle{ruled}{labelfont=normalfont,labelsep=colon,strut=off}
\floatstyle{ruled}
\newfloat{listing}{tb}{lst}{}
\floatname{listing}{Listing}

\usepackage{booktabs}
\usepackage{amsmath}
\usepackage{amssymb}
\usepackage{multirow}
\usepackage{xcolor}
\usepackage{array}
\usepackage{colortbl}
\usepackage{tcolorbox}
\usepackage{fvextra}
\tcbuselibrary{breakable}
\tcbset{breakable}
\RecustomVerbatimEnvironment{verbatim}{Verbatim}{breaklines=true,breakanywhere=true}

\title{MuEvo: LLM-Driven Evolution of Multi-Heuristic Ensemble}

\author{
  Haoze Lv\textsuperscript{\rm 1}\equalcontrib,
  Ning Lu\textsuperscript{\rm 1,\rm 2}\equalcontrib,
  Shengcai Liu\textsuperscript{\rm 1}\corresponding
  Shaofeng Zhang\textsuperscript{\rm 1},
  Ke Tang\textsuperscript{\rm 1}
}
\affiliations{
  \textsuperscript{\rm 1}Guangdong Provincial Key Laboratory of Brain-Inspired Intelligent Computation,\\
  Department of Computer Science and Engineering, Southern University of Science and Technology,\\
  Shenzhen 518055, China\\
  \textsuperscript{\rm 2}The Hong Kong University of Science and Technology\\
  liusc3@sustech.edu.cn
}

\begin{document}

\maketitle

\begin{abstract}
Large language model-based automated heuristic design (LLM-AHD) has shown strong potential in discovering effective heuristics for combinatorial optimization problems. However, existing methods primarily optimize a single heuristic, whereas practical optimization frameworks often rely on multiple interacting components. Directly extending single-heuristic methods is challenging because early component selection can overlook components with late potential, while independent evolution ignores inter-component dependencies. We propose MuEvo, an LLM-driven framework for evolving heuristic ensembles under ensemble-level feedback. MuEvo combines Dynamic Component Management, which uses short-budget probing and a reversible lifecycle to revise component priorities throughout the search, with LLM-Driven Co-Evolution, which coordinates component populations through Multi-Ensemble Evaluation, Cross-Component Information Sharing, Relation-Guided Pair Evolution, and Adaptive Budget Allocation. We evaluate MuEvo on selection hyper-heuristics and componentized ant colony optimization across four combinatorial optimization domains. Results show that MuEvo consistently improves human-designed frameworks and outperforms representative multi-component extensions of state-of-the-art LLM-AHD methods, demonstrating its effectiveness across both controller-mediated heuristic pools and functionally differentiated algorithmic components.
\end{abstract}

\input{Sections/1_Introduction.tex}

\input{Sections/2_Relatework.tex}
\input{Sections/3_Method.tex}
\input{Sections/4_Experiment.tex}

\input{Sections/5_Conclusion.tex}

\bibliography{aaai2027}

\clearpage
\setcounter{secnumdepth}{2}
\input{Sections/6_Appendix.tex}

\end{document}

%% file: Sections/1_Introduction.tex
\section{Introduction}
\label{sec:intro}

Given the wide applications of combinatorial optimization problems (COPs) in real world, designing effective heuristics has received significant attention~\cite{TSPAP:matai2010traveling, FlowshopAP:rajendran1993heuristic}. 
However, the traditional manual design process relies heavily on domain knowledge and involves costly trial-and-error loops~\cite{Intro1COPAppliaction:desale2015heuristic}. 
To address these issues, automated heuristic design (AHD) has been studied as a promising approach to simplify the design process~\cite{HHSurvey2010:burke2010classification}. 

Traditional AHD approaches, such as genetic programming, automatically design heuristics by searching in a predefined space~\cite{HH8:langdon2013foundations, GPAP1:mei2022explainable}. 
However, the predefined search spaces and operator sets still rely on human knowledge~\cite{EOH:liu2024evolution}.
Recent studies have shown that large language models (LLMs) offer new approaches for AHD, demonstrating superior performance with minimal reliance on expert knowledge~\citep{FunSearch:romeraparedes2024mathematical,EOH:liu2024evolution,AlphaEvolve:journals/corr/abs-2506-13131}. They maintain a population of heuristic code snippets and employ LLMs as crossover or mutation operators to generate new variants~\citep{FunSearch:romeraparedes2024mathematical}.

\begin{figure}[t!]
\centering
\begin{minipage}{0.49\columnwidth}
    \centering
    \includegraphics[width=\linewidth]{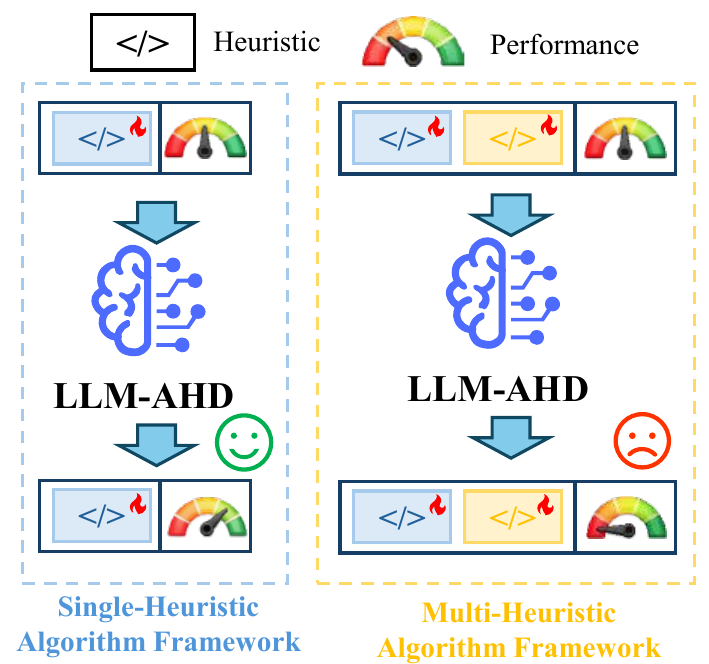}
    \par
    \textbf{(a)}
\end{minipage}
\hfill
\begin{minipage}{0.49\columnwidth}
    \centering
    \includegraphics[width=\linewidth]{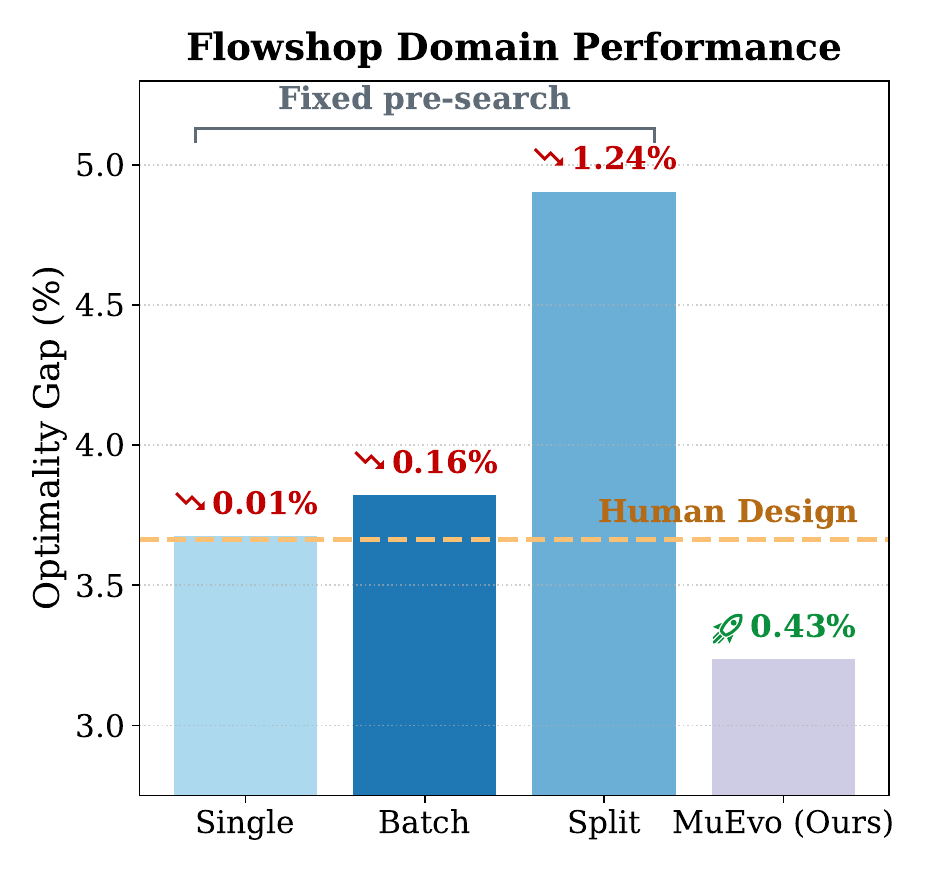}
    \par
    \textbf{(b)}
\end{minipage}
\caption{\textbf{Existing LLM-AHD methods do not explicitly support dynamic coadaptation in multi-heuristic frameworks.} (a) Existing methods primarily design a single heuristic, whereas practical optimization frameworks often rely on multiple interacting heuristics that jointly determine the search behavior. (b) Performance of ReEvo-based Single, Batch, and Split extensions under fixed pre-search, compared with Human Design and MuEvo on Flowshop Problem. Lower values indicate better performance.}
\label{fig:combined_intro_comparison}
\vspace{-5mm}
\end{figure}

However, existing LLM-AHD methods primarily optimize a single heuristic~\citep{FunSearch:romeraparedes2024mathematical,ReEvo:ye2024reevo}, whereas practical optimization frameworks often rely on multiple functionally distinct heuristics that jointly determine search behavior. For example, memetic algorithms may employ separate heuristics for initialization, crossover, and local search~\cite{SAHiD:tang2009memetic,HHApplication:de2021comparative}. We refer to a collection of such jointly operating heuristics within an algorithmic framework as a \textit{heuristic ensemble}.
Recent work has begun to extend LLM-AHD to multiple components, for example through fixed pre-search that retains a subset for subsequent evolution~\citep{AutoSAT:sun2025automatically}. This strategy faces two limitations: components evaluated only in the initial ensemble may be underestimated, and evolving the retained components without explicit coordination may overlook their dependencies. As illustrated in Figure~1, our controlled ReEvo-based Single, Batch, and Split extensions show inconsistent performance under fixed pre-search, whereas MuEvo achieves the lowest optimality gap.

To address these challenges, we propose \textbf{MuEvo}, an LLM-driven method for evolving heuristic ensembles in multi-heuristic algorithm frameworks. MuEvo integrates Dynamic Component Management, which uses short-budget probing and a feedback-driven lifecycle to allow initially underestimated components to re-enter the search, with LLM-Driven Co-Evolution, which maintains component-wise populations and coordinates their evolution through ensemble-aware evaluation, cross-component collaboration, and adaptive budget allocation. Together, these mechanisms transform one-shot component selection and isolated optimization into a dynamic co-evolutionary process driven by ensemble-level feedback.

We evaluate MuEvo on two multi-component frameworks with distinct interaction structures: selection hyper-heuristics (SHHs), where a high-level controller dynamically invokes low-level heuristics (LLHs) from a largely peer-level pool~\cite{SHHRL:de2020hyper,HH11:qin2021novel}, and componentized ant colony optimization (ACO), where components at different workflow stages exhibit structured dependencies~\cite{ACO:dorigo1996ant}. Across four COP domains, MuEvo generates competitive heuristic ensembles under both interaction structures, reducing the optimality gap of the strongest LLM-AHD baseline from 6.26\% to 5.24\% on TSP and from 3.51\% to 3.24\% on Flowshop.

Our main contributions are summarized as follows:
\begin{itemize}
  \item We systematically propose the problem of multi-heuristic LLM-AHD, highlighting the importance and challenges of evolving multiple heuristics.
  \item We propose MuEvo, which combines Dynamic Component Management with LLM-Driven Co-Evolution to revise component priorities using subsequent ensemble feedback and promote coadaptation across component populations.
  \item We conduct comprehensive experiments across SHH and componentized ACO frameworks covering four combinatorial optimization domains, demonstrating strong performance as well as generalization across high-level controllers and LLM backbones.
\end{itemize}

%% file: Sections/2_Relatework.tex
\section{Preliminaries and Related Work}
\label{sec:related_work}

\subsection{Problem Definition}
\label{sec:problem_definition}

Let $P$ be the COP of interest, and let $\mathcal{D}$ denote the distribution over its instances.
For each component position $i$, let $\mathbb{C}_i$ represent the space of syntactically correct and executable heuristic code compatible with that position.
We define $\mathcal{F}$ as an algorithm framework (e.g., an SHH framework).
When instantiated with an ordered heuristic ensemble
$\mathcal{E}=(h_1,\ldots,h_N)\in\prod_{i=1}^{N}\mathbb{C}_i$,
where each $h_i\in\mathbb{C}_i$, the framework solves an instance
$I\sim\mathcal{D}$ to produce a feasible solution $y$.

\textbf{Multi-Heuristic AHD.}
The objective of Multi-Heuristic AHD is to identify an optimal heuristic ensemble $\mathcal{E}^*$ that minimizes the expected objective value over the distribution $\mathcal{D}$:
\begin{equation}
\mathcal{E}^*
=
\arg\min_{\mathcal{E}\in\prod_{i=1}^{N}\mathbb{C}_i}
\mathbb{E}_{I\sim\mathcal{D}}
\!\left[
Q(\mathcal{F}(\mathcal{E}),I)
\right],
\end{equation}
where $Q(\mathcal{F}(\mathcal{E}),I)$ quantifies the objective value of the solution generated by the framework instantiated with $\mathcal{E}$ on instance $I$, with lower values indicating better performance.

\textbf{Single-Heuristic AHD as a Special Case.}
Single-heuristic AHD corresponds to the special case $N=1$.
Let $\mathcal{E}=(h)$ with $h\in\mathbb{C}_1$; then Multi-Heuristic AHD reduces to
\begin{equation}
h^*
=
\arg\min_{h\in\mathbb{C}_1}
\mathbb{E}_{I\sim\mathcal{D}}
\!\left[
Q(\mathcal{F}((h)),I)
\right],
\end{equation}
which is exactly the single-heuristic AHD objective.
This formulation highlights that Multi-Heuristic AHD searches over
the joint component space $\prod_{i=1}^{N}\mathbb{C}_i$, which reduces
to $\mathbb{C}^{N}$ when all components share the same code space.
More importantly, the utility of each component depends on its
collaborators, making the ensemble objective generally non-separable.

\subsection{LLM-AHD}
LLM-based AHD treats LLMs as search operators for generating executable heuristics, with solver performance guiding iterative refinement. FunSearch and EoH established this paradigm, while ReEvo, MCTS-AHD, and PathWise extended it with reflection, tree search, and model-based planning, respectively~\cite{FunSearch:romeraparedes2024mathematical,EOH:liu2024evolution,ReEvo:ye2024reevo,MCTS-AHD:zheng2025monte,PathWise:gungordu2026pathwise,E2OC:qiu2026evolving}. It has since been applied to scheduling, MILP, and SAT~\cite{LLMFS:li2025llm,DSEvo:zhang2025dhevo,SAT:chen2025dasathco}.
While most LLM-AHD methods remain centered on a single heuristic, several recent studies have begun to extend this paradigm to multiple heuristics or components, albeit with limited support for their coadaptation. VRPAgent jointly designs two routing operators but does not maintain component-wise populations~\cite{VRPAgent:hottung2025vrpagent}. AutoModSAT uses fixed pre-search to select SAT-solver functions and restricts subsequent evolution to the retained subset, without explicitly modeling their coadaptation~\citep{AutoSAT:sun2025automatically}. EoH-S constructs a portfolio of separate heuristic--framework pairs~\citep{EOH-S:liu2025eoh}. CoEvo-AHD co-evolves two component-specific operator populations for bi-component coupled optimization and evaluates their interaction under the full problem objective, but its formulation is designed for two-component problem structures~\citep{CoEvo-HH:kuang2026llm}. In contrast, MuEvo targets broader multi-component algorithm frameworks and co-evolves multiple interdependent heuristics through component-specific populations.

\begin{figure}[t]
    \centering
    \includegraphics[width=\linewidth]{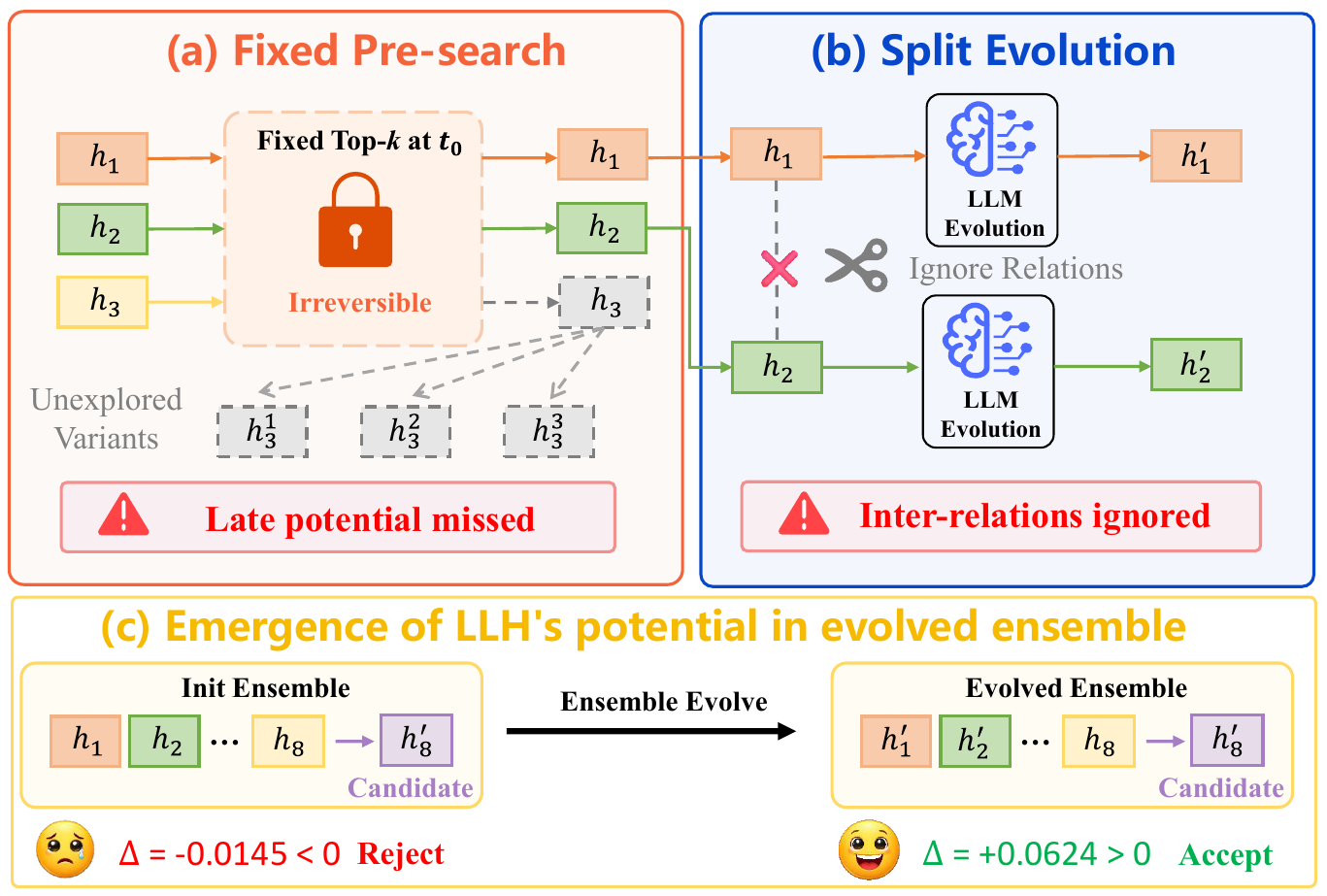}
    \caption{\textbf{Limitations of fixed selection and independent evolution.} Fixed pre-search may miss late potential, while independent evolution ignores component dependencies. In TSP-SHH, the same \(h_8'\) candidate degrades the initial ensemble but improves the evolved one.}
    \label{fig:motivation}
    \vspace{-2mm}
\end{figure}


\subsection{Multi-Heuristic Frameworks}

We study multi-heuristic LLM-AHD in two structurally distinct frameworks: selection hyper-heuristics (SHHs) and componentized ant colony optimization (ACO). In SHHs, a high-level controller adaptively selects low-level heuristics (LLHs) from a predefined pool~\cite{HH1:burke2013hyper,HH2:drake2020recent}. Our SHH testbed follows the CHeSC 2011 HyFlex interface, with domain-specific LLH pools and multiple high-level controllers~\cite{Hyflex:ochoa2012hyflex}. In componentized ACO, the classical Ant System uses pheromone trails and problem-dependent heuristic information to guide stochastic solution construction, with all ants reinforcing their solutions~\cite{ACO:dorigo1996ant}. Nine replaceable components incorporate mechanisms from later ACO variants, including ACS-style candidate lists and local pheromone updates, rank-based and best-so-far reinforcement, stagnation control, and elite-solution memory~\cite{ACO:dorigo1997acs,ACO:bullnheimer1999rank,ACO:stutzle2000maxmin,ACO:guntsch2002population}. Together, the two frameworks capture controller-mediated interactions among peer LLHs and structured dependencies across algorithmic stages. Complete definitions and implementation details for the SHH and ACO frameworks are provided in Appendices A and B, respectively.

%% file: Sections/3_Method.tex
\section{Methodology}
\label{sec:method}

\begin{figure*}[t]
\centering
\resizebox{0.95\textwidth}{!}{\includegraphics{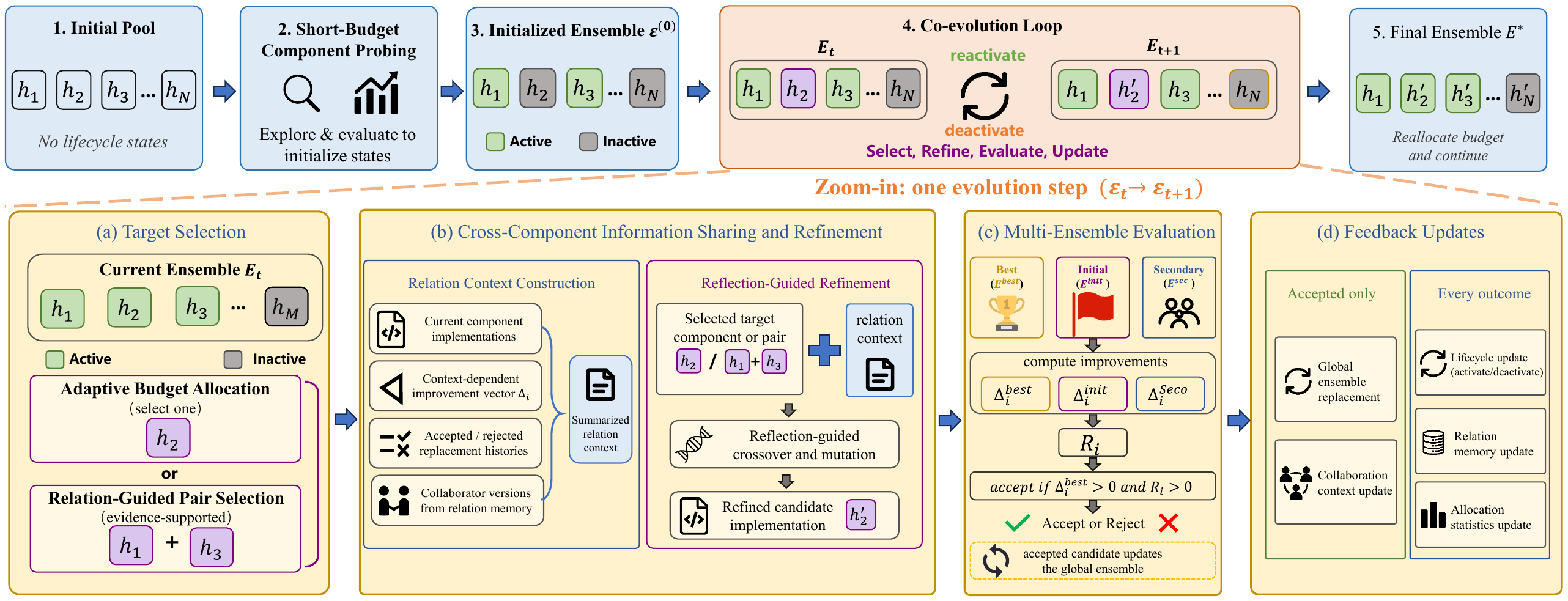}}
\caption{\textbf{Overview of MuEvo.} Dynamic Component Management retains all valid components in a reversible lifecycle, while LLM-Driven Co-Evolution combines Multi-Ensemble Evaluation, Cross-Component Information Sharing, Relation-Guided Pair Evolution, and Adaptive Budget Allocation.}
\label{fig:framework_workflow}
\vspace{-1em}
\end{figure*}

In this section, we present \textbf{MuEvo} to address two key challenges in multi-heuristic design: the premature exclusion of promising components and overlooked inter-component dependencies. As illustrated in Figure~\ref{fig:framework_workflow}, MuEvo addresses these challenges by combining Dynamic Component Management with LLM-Driven Co-Evolution. We first motivate this design and then detail the two core mechanisms of MuEvo.

\subsection{Motivation and Method Overview}
\label{subsec:motivation}

Recent multi-component LLM-AHD work uses fixed pre-search to retain a subset of components for later optimization~\citep{AutoSAT:sun2025automatically}. A natural extension is then to evolve the retained components independently while freezing the rest, yielding the two-step pipeline shown in Figure~\ref{fig:motivation}(a--b). Despite its efficiency, this pipeline has two limitations. \textbf{Late potential missed.} Components are ranked only in the initial ensemble, although their utility may change as their collaborators evolve; irreversible Top-$k$ selection can therefore discard components whose potential emerges later. \textbf{Inter-component dependencies ignored.} Independent evolution neglects synergy, redundancy, and conflict among components, so individually improved components may still form a weak ensemble. Figure~\ref{fig:motivation}(c) provides a concrete observed case from TSP-SHH: the same \(h_8'\) candidate increases the gap by \(0.0145\) percentage points in the initial ensemble, but reduces it by \(0.0624\) percentage points after the ensemble evolves. This sign reversal shows that a component update judged unpromising in the initial ensemble may exhibit potential later, while its utility varies with the state of the ensemble, providing direct evidence of inter-LLH dependencies and supporting the two motivations above.

MuEvo addresses these limitations through Dynamic Component Management and LLM-Driven Co-Evolution. The former combines short-budget component probing with a reversible lifecycle to revise component priorities as new evidence emerges, while the latter coordinates component populations through Multi-Ensemble Evaluation, Cross-Component Information Sharing, Relation-Guided Pair Evolution, and Adaptive Budget Allocation. The following two subsections detail these mechanisms.

\subsection{Dynamic Component Management}
Fixed pre-search converts an early ranking from the initial ensemble into an irreversible Top-$k$ decision, even though component utility may change as the ensemble evolves. MuEvo instead combines short-budget component probing with a feedback-driven lifecycle to preserve such late potential. As summarized in Algorithm~\ref{alg:muevo}, it probes all valid components, initializes them as Active or Inactive (lines~3-4), and updates their states and evolution priorities from subsequent co-evolution outcomes (line~19), allowing underestimated components to re-enter the search.

\subsubsection{Short-budget component probing.}
To estimate initial improvement potential, MuEvo briefly evolves each component under the same probing budget while holding its collaborators fixed. Let $f_i^0$ denote the ensemble fitness before probing $h_i$ and $f_i^*$ the best fitness observed during the probe. Since lower fitness is better, the probing score is defined as
\begin{equation}
S_i^{\mathrm{probe}}=f_i^0-f_i^*.
\label{eq:probe_score}
\end{equation}
A larger score indicates greater improvement potential under the fixed probing budget.

\subsubsection{Dynamic component lifecycle.}
Probe scores provide only initial evidence and therefore determine reversible priorities rather than permanent eligibility. MuEvo ranks all valid components by $S_i^{\mathrm{probe}}$, initializes the top $K_A$ as Active and the remainder as Inactive, and reserves regular evolution for Active components while retaining Inactive ones for later reactivation. 
State transitions are driven by ensemble-level feedback. An Active component with $\tau_{\mathrm{reject}}$ consecutive Multi-Ensemble Evaluation rejections is deactivated for $L_{\mathrm{cool}}$ rounds, provided that at least one other Active component remains, and cannot be selected or reactivated during cooldown. When global stagnation reaches $\tau_{\mathrm{stag}}$ and at least $\tau_{\mathrm{int}}$ rounds have elapsed since the previous activation, the highest-priority non-cooled Inactive component is promoted to Active and force-selected for one round. Activation priority favors fewer evaluation rejections, higher probing scores, larger historical replacement gains, and fewer prior activations. An accepted replacement clears the component's rejection and cooldown records and resets global stagnation.

\subsection{LLM-Driven Co-Evolution}
\label{subsec:main_loop}

LLHs in a multi-heuristic system are often interdependent, so optimizing one in isolation may weaken the resulting ensemble. Existing LLM-AHD methods provide limited support for such component-level coadaptation. Cooperative coevolution offers a natural formulation by decomposing the system into subcomponents, maintaining a population for each, and coordinating their evolution to promote coadaptation~\citep{CC:yang2008large}.
MuEvo instantiates this formulation by treating each LLH as a subcomponent and maintaining a dedicated population for it. The within-population evolutionary process follows the reflective evolution design of ReEvo~\citep{ReEvo:ye2024reevo}, employing reflection-guided crossover and mutation to refine heuristic code.  
As summarized in Algorithm~\ref{alg:muevo}, Adaptive Budget Allocation selects an Active LLH for evolution, while Relation-Guided Pair Evolution periodically selects an evidence-supported pair (line~10). Cross-Component Information Sharing incorporates relation evidence into the evolutionary context (line~11), and Multi-Ensemble Evaluation determines whether the resulting candidate enters the global ensemble (line~13). An accepted replacement updates the global ensemble and collaboration contexts (line~16-17), while every evaluation outcome updates the component lifecycle and relation memory (line~19), forming a closed loop in which evidence from one population guides the subsequent evolution of others. The following subsections detail these mechanisms.

\subsubsection{Multi-Ensemble Evaluation.}
A candidate component must be evaluated within a complete ensemble, yet a single collaborator configuration may favor context-specific improvements. MuEvo therefore evaluates each candidate in three representative collaboration contexts: the current Best ensemble, the fixed Initial ensemble, and a Secondary ensemble selected from the archive based on quality and diversity. To keep the evaluation cost fixed, the same per-candidate solver-run budget is shared across the three contexts, with the detailed allocation provided in Appendix C. After within-population evolution proposes $h_i'$, its improvement in context $k$ is defined as
\begin{equation}
\Delta_i^k=f(\mathcal{E}_k)-f(\mathcal{E}_k[h_i\leftarrow h_i']),
\end{equation}
and aggregated as
\begin{equation}
R_i=\sum_k w_k\Delta_i^k+\lambda_{\min}\min_k\Delta_i^k-\lambda_{\mathrm{var}}\operatorname{Std}(\{\Delta_i^k\}).
\label{eq:multi_ensemble_score}
\end{equation}
The three terms capture weighted improvement, worst-context performance, and cross-context consistency, respectively. A candidate is accepted only if $\Delta_i^{\mathrm{Best}}>0$ and $R_i>0$. Jointly evolved pairs are evaluated analogously, with both replacements inserted into each context and accepted or rejected together. The context-dependent improvement vector $\boldsymbol{\Delta}_i=[\Delta_i^{\mathrm{Best}},\Delta_i^{\mathrm{Initial}},\Delta_i^{\mathrm{Secondary}}]$ is retained as evidence for subsequent co-evolution.

\subsubsection{Cross-Component Information Sharing.}
Components may be synergistic, redundant, or conflicting, so evolving a target component requires information about the rest of the ensemble. Cross-Component Information Sharing constructs this context from current implementations and multi-ensemble feedback. Before refining $h_i$, MuEvo summarizes the functional roles and dependencies of the current components, while relation memory stores, for each pair, inferred dependencies, collaborator versions associated with accepted and rejected replacements, and historical context-dependent gains. Together with $\boldsymbol{\Delta}_i$, these records form a target-specific relation context that distinguishes robust from collaborator-specific gains and conditions the LLM-based refinement of $h_i$, allowing evidence from one component population to guide the evolution of others.

\subsubsection{Relation-Guided Pair Evolution.}
To further promote coadaptation among interacting components, Relation-Guided Pair Evolution jointly refines an evidence-supported pair as a cooperative unit. Every $q$ evolution rounds, provided sufficient evaluation budget remains, MuEvo ranks eligible component pairs by combining structural dependency inferred from their current implementations with empirical interaction evidence stored in relation memory. The highest-ranked pair is jointly refined and evaluated as a single candidate, with the detailed scoring and tie-breaking rules provided in Appendix C. The resulting pair is evaluated as a single joint candidate through Multi-Ensemble Evaluation, so both components are accepted or rejected together.

\subsubsection{Adaptive Budget Allocation.}
\label{subsubsec:meta_controller}

Co-evolution must determine which Active component should receive the next regular evaluation. Adaptive Budget Allocation ranks eligible components outside cooldown according to their current improvement potential while discouraging persistent over-allocation. Except when a reactivation event force-selects a newly activated component, MuEvo assigns each eligible $h_i$ the score
\begin{equation}
P_i=w_I I_i+w_V\widetilde V_i+w_S\min\left(\frac{z_i}{H_z},1\right)+w_R\widetilde r_i-w_\rho\rho_i.
\label{eq:improvement_potential}
\end{equation}
Here, $I_i$ denotes the recent improvement rate derived from verified population fitness, $\widetilde V_i$ the normalized fitness variance, and $\widetilde r_i$ the normalized mean ensemble-replacement reward over the most recent $H_r$ selections; $z_i$ counts consecutive selections since the last accepted replacement, $H_z$ controls saturation of the stagnation bonus, and $\rho_i$ is the fraction of regular evolution rounds allocated to $h_i$. These terms jointly capture recent progress, population diversity, temporary stagnation, ensemble-level reward, and allocation frequency. MuEvo selects the component with the largest $P_i$, refines it through reflection-guided crossover and mutation, and updates the score statistics, component lifecycle, relation memory, and ensemble archive from the replacement outcome. Scheduled Relation-Guided Pair Evolution follows its own pair-selection rule and consumes the same evaluation budget.

\begin{algorithm}[t]
\caption{MuEvo Overview}
\label{alg:muevo}
\begin{algorithmic}[1]
\STATE \textbf{Input:} Framework $\mathcal{F}$, initial ensemble $E_0$, probe budget $B_{\mathrm{probe}}$, evolution budget $B_{\mathrm{evo}}$
\STATE \textbf{Output:} Optimized ensemble $E^*$
\STATE $\mathcal{P} \leftarrow \textsc{ProbeComponents}(\mathcal{F},E_0,B_{\mathrm{probe}})$
\STATE $(\mathcal{A},\mathcal{I}) \leftarrow \textsc{InitializeLifecycle}(\mathcal{P})$
\STATE $E^* \leftarrow E_0$
\STATE $(\mathcal{M},\mathcal{C}) \leftarrow \textsc{InitializeSearchState}(E_0)$
\STATE $b \leftarrow 0,\; t \leftarrow 0$
\WHILE{$b < B_{\mathrm{evo}}$}
\STATE $t \leftarrow t+1$
\STATE $J_t \leftarrow \textsc{SelectTargetOrPair}(\mathcal{A},\mathcal{I},\mathcal{M},t)$
\STATE $\Gamma_t \leftarrow \textsc{BuildRelationContext}(J_t,E^*,\mathcal{M})$
\STATE $h'_{J_t} \leftarrow \textsc{Evolve}(h_{J_t},\Gamma_t)$
\STATE $g_t \leftarrow \textsc{MultiEnsembleEvaluation}(E^*,J_t,h'_{J_t},\mathcal{C})$
\STATE $b \leftarrow b+1$
\IF{$\textsc{Accept}(g_t)$}
\STATE $E^* \leftarrow \textsc{Replace}(E^*,J_t,h'_{J_t})$
\STATE $\mathcal{C} \leftarrow \textsc{UpdateContexts}(\mathcal{C},E^*)$
\ENDIF
\STATE $\textsc{UpdateLifecycleAndMemory}(J_t,g_t,\mathcal{A},\mathcal{I},\mathcal{M})$
\ENDWHILE
\RETURN $E^*$
\end{algorithmic}
\end{algorithm}

%% file: Sections/4_Experiment.tex
\begin{table*}[ht]
\resizebox{\textwidth}{!}{%
\begin{tabular}{l|cccc|cccc}
\hline
\textbf{Domain} & \multicolumn{4}{c|}{\textbf{TSP}} & \multicolumn{4}{c}{\textbf{CVRP}} \\
\hline
\textbf{Dataset} & \textbf{TSP-SS} & \textbf{TSP-S} & \textbf{TSP-M} & \textbf{TSP-L} & \textbf{Set-A} & \textbf{Set-B} & \textbf{Set-P} & \textbf{CMT} \\
\hline
\textbf{Method} & Cost (Gap) $\downarrow$ & Cost (Gap) $\downarrow$ & Cost (Gap) $\downarrow$ & Cost (Gap) $\downarrow$ & Cost (Gap) $\downarrow$ & Cost (Gap) $\downarrow$ & Cost (Gap) $\downarrow$ & Cost (Gap) $\downarrow$ \\
\hline
Default-SHH & 30211.7 (1.875\%) & 26504.9 (4.756\%) & 92521.4 (5.765\%) & 44680.6 (7.157\%) & 1175.19 (12.26\%) & 1053.06 (8.871\%) & 645.892 (9.817\%) & 1051.51 (8.649\%) \\
POMO & 35100.4 (21.44\%) & 37924.0 (53.78\%) & 158371 (84.92\%) & 64956.8 (110.2\%) & 1278.48 (24.70\%) & 1157.57 (20.03\%) & 791.792 (43.43\%) & 1136.62 (16.73\%) \\
\hline
EoH-Single & 30234.5 (2.032\%) & 26577.5 (4.964\%) & 93193.8 (7.194\%) & 40316.6 (17.73\%) & 1177.4 (12.28\%) & 1050.93 (8.618\%) & 644.467 (9.116\%) & 1078.24 (10.86\%) \\
ReEvo-Single & 30000.4 (0.919\%) & 26136.6 (3.584\%) & 91530.6 (4.712\%) & 44542.0 (6.736\%) & 1151.28 (9.950\%) & 1051.75 (8.633\%) & 638.179 (8.185\%) & 1061.09 (9.197\%) \\
MCTS-AHD-Single & 29957.7 (0.723\%) & 26033.2 (2.988\%) & 91524.4 (4.504\%) & 44462.6 (6.505\%) & 1147.98 (9.124\%) & 1033.39 (6.583\%) & 629.938 (6.084\%) & 1083.44 (11.11\%) \\
\hline
EoH-Batch & 30131.9 (1.532\%) & 26456.9 (4.663\%) & 92488.1 (5.992\%) & 45195.9 (12.55\%) & 1090.99 (4.136\%) & 1010.56 (4.610\%) & 616.367 (4.099\%) & \textbf{999.800 (2.730\%)} \\
ReEvo-Batch & 30173.9 (1.700\%) & 26399.8 (4.390\%) & 92590.5 (5.675\%) & 44766.5 (7.865\%) & 1148.23 (9.644\%) & 1046.10 (8.067\%) & 632.783 (7.277\%) & 1050.59 (7.962\%) \\
MCTS-AHD-Batch & \textbf{29925.8 (0.594\%)} & 25987.6 (2.954\%) & 91083.4 (4.308\%) & 44041.8 (6.261\%) & 1143.84 (8.714\%) & 1028.30 (5.998\%) & 630.763 (6.051\%) & 1084.24 (11.12\%) \\
\hline
EoH-Split & 30078.5 (1.316\%) & 26427.7 (4.381\%) & 92343.7 (6.044\%) & 40344.4 (18.21\%) & 1143.05 (8.647\%) & 1025.14 (5.747\%) & 625.350 (5.834\%) & 1058.51 (8.522\%) \\
ReEvo-Split & 29999.5 (0.832\%) & 26902.8 (6.577\%) & 100873 (16.40\%) & 46035.1 (14.64\%) & 1126.87 (7.627\%) & 1019.59 (5.590\%) & 622.092 (5.605\%) & 1027.11 (5.939\%) \\
MCTS-AHD-Split & 30653.2 (3.182\%) & 26700.2 (5.798\%) & 90057.4 (15.63\%) & 41966.0 (16.45\%) & 1087.38 (4.059\%) & 1000.03 (3.592\%) & 609.308 (3.968\%) & 1046.70 (7.332\%) \\
\hline
\rowcolor{gray!15} \textbf{MuEvo} & 29935.8 (0.651\%) & \textbf{25815.0 (2.141\%)} & \textbf{90672.5 (3.657\%)} & \textbf{43909.0 (5.237\%)} & \textbf{1057.14 (1.387\%)} & \textbf{985.174 (2.135\%)} & \textbf{603.870 (2.300\%)} & 949.127 (4.213\%) \\
\hline\hline
\textbf{Domain} & \multicolumn{4}{c|}{\textbf{BPP}} & \multicolumn{4}{c}{\textbf{Flowshop}} \\
\hline
\textbf{Dataset} & \textbf{Falkenauer-U} & \textbf{Falkenauer-T} & \textbf{Scholl-3} & \textbf{Schwerin} & \textbf{VRF20} & \textbf{VRF40} & \textbf{VRF60} & \textbf{VRF100} \\
\hline
\textbf{Method} & Bins (Gap) $\downarrow$ & Bins (Gap) $\downarrow$ & Bins (Gap) $\downarrow$ & Bins (Gap) $\downarrow$ & Makespan (Gap) $\downarrow$ & Makespan (Gap) $\downarrow$ & Makespan (Gap) $\downarrow$ & Makespan (Gap) $\downarrow$ \\
\hline
Default-SHH & 190.667 (0.754\%) & 82.9958 (6.484\%) & 58.1800 (3.530\%) & 20.3467 (2.166\%) & 1724.40 (0.173\%) & 2759.63 (1.585\%) & 3796.15 (1.858\%) & 8200.53 (3.663\%) \\
\hline
EoH-Single & 190.367 (0.592\%) & 79.2250 (2.837\%) & 57.0400 (1.501\%) & 20.1133 (0.963\%) & 1724.69 (0.199\%) & 2753.01 (1.358\%) & 3800.1 (1.955\%) & 8278.57 (4.633\%) \\
ReEvo-Single & 189.688 (0.299\%) & 79.1542 (2.822\%) & 56.5100 (0.556\%) & 20.3967 (2.387\%) & 1724.31 (0.170\%) & 2759.60 (1.584\%) & 3796.06 (1.856\%) & 8201.28 (3.676\%) \\
MCTS-AHD-Single & 189.996 (0.455\%) & 81.4208 (5.009\%) & 58.1700 (3.512\%) & 20.4500 (2.686\%) & 1722.25 (0.058\%) & 2745.26 (1.087\%) & 3780.26 (1.452\%) & 8234.87 (4.095\%) \\
\hline
EoH-Batch & 189.867 (0.454\%) & 78.4833 (2.321\%) & 56.6200 (0.749\%) & 20.1933 (1.362\%) & 1724.98 (0.211\%) & 2760.53 (1.610\%) & 3800.10 (1.960\%) & 8262.31 (4.449\%) \\
ReEvo-Batch & 189.146 (0.102\%) & 78.5000 (2.326\%) & 56.4500 (0.450\%) & 20.1667 (1.245\%) & 1722.71 (0.080\%) & 2744.34 (1.051\%) & 3777.29 (1.377\%) & 8212.45 (3.822\%) \\
MCTS-AHD-Batch & 189.121 (0.072\%) & 78.6375 (2.408\%) & 57.7100 (2.692\%) & 20.1934 (1.382\%) & 1721.98 (0.046\%) & 2742.04 (0.971\%) & 3777.56 (1.381\%) & 8187.62 (3.506\%) \\
\hline
EoH-Split & 189.158 (0.097\%) & 78.5000 (2.326\%) & 56.4700 (0.485\%) & 20.1133 (0.958\%) & 1722.98 (0.099\%) & 2749.81 (1.240\%) & 3813.03 (2.272\%) & 8302.73 (5.004\%) \\
ReEvo-Split & 190.658 (0.771\%) & 83.5833 (7.073\%) & 58.3200 (3.779\%) & 20.3667 (2.267\%) & 1723.85 (0.145\%) & 2748.69 (1.210\%) & 3776.13 (1.705\%) & 8174.62 (4.904\%) \\
MCTS-AHD-Split & 190.575 (0.706\%) & 82.3000 (5.757\%) & 57.9800 (3.175\%) & 20.4067 (2.482\%) & 1722.64 (0.079\%) & 2748.21 (1.184\%) & 3783.91 (1.542\%) & 8290.35 (4.778\%) \\
\hline
\rowcolor{gray!15} \textbf{MuEvo} & \textbf{189.017 (0.051\%)} & \textbf{78.4667 (2.159\%)} & \textbf{56.4000 (0.360\%)} & \textbf{20.0067 (0.394\%)} & \textbf{1721.54 (0.022\%)} & \textbf{2740.78 (0.938\%)} & \textbf{3771.57 (1.235\%)} & \textbf{8169.21 (3.236\%)} \\
\hline
\end{tabular}%
}
\centering
\caption{\textbf{Performance comparison across four problem domains.} Values report mean cost/bins/makespan and mean gap (\%). Cost and Gap are averaged over 3 independent runs. For all metrics, lower is better. }
\label{tab:main_results}
\end{table*}

\begin{table*}[t]
\resizebox{\textwidth}{!}{%
\begin{tabular}{l|cccc|cccc}
\hline
\textbf{Domain} & \multicolumn{4}{c|}{\textbf{TSP}} & \multicolumn{4}{c}{\textbf{CVRP}} \\
\hline
\textbf{Dataset} & \textbf{TSP-SS} & \textbf{TSP-S} & \textbf{TSP-M} & \textbf{TSP-L} & \textbf{Set-A} & \textbf{Set-B} & \textbf{Set-P} & \textbf{CMT} \\
\hline
\textbf{Method} & Cost (Gap) $\downarrow$ & Cost (Gap) $\downarrow$ & Cost (Gap) $\downarrow$ & Cost (Gap) $\downarrow$ & Cost (Gap) $\downarrow$ & Cost (Gap) $\downarrow$ & Cost (Gap) $\downarrow$ & Cost (Gap) $\downarrow$ \\
\hline
Default ACO & 31062.8 (4.830\%) & 37285.0 (65.96\%) & 239798 (173.6\%) & 126947 (199.8\%) & 1224.73 (17.51\%) & 1054.83 (9.589\%) & 693.292 (16.36\%) & 1105.09 (14.83\%) \\
DeepACO & 33417.6 (14.58\%) & 36788.3 (43.72\%) & 176989 (105.1\%) & 88034.6 (205.2\%) & 2066.86 (96.97\%) & 1796.38 (83.81\%) & 922.472 (45.65\%) & 1241.51 (26.81\%) \\
\hline
EoH-Single & 31679.7 (6.759\%) & 29360.2 (19.24\%) & 130805 (55.58\%) & 99687.8 (158.1\%) & 1086.55 (3.834\%) & 991.212 (2.804\%) & 610.983 (3.561\%) & 1023.63 (5.024\%) \\
ReEvo-Single & 30851.1 (3.754\%) & 29273.9 (17.89\%) & 134035 (54.51\%) & 65091.3 (55.53\%) & 1084.46 (3.689\%) & \textbf{988.339 (2.530\%)} & 609.597 (3.342\%) & 1045.62 (6.768\%) \\
MCTS-AHD-Single & 30763.3 (3.754\%) & 30280.0 (26.23\%) & 208238 (138.4\%) & 113967 (174.0\%) & 1110.81 (6.097\%) & 1001.54 (3.859\%) & 614.353 (4.161\%) & 1014.24 (4.697\%) \\
\hline
EoH-Batch & 30416.1 (1.893\%) & 28853.5 (16.08\%) & 120525 (39.89\%) & 58273.3 (39.65\%) & 1109.85 (6.071\%) & 998.067 (3.628\%) & 623.633 (5.515\%) & 1027.81 (5.951\%) \\
ReEvo-Batch & 30467.9 (2.342\%) & 28236.2 (13.21\%) & 114530 (29.11\%) & 72177.4 (72.10\%) & 1085.48 (3.901\%) & 988.136 (2.610\%) & 611.561 (3.582\%) & 998.419 (2.970\%) \\
MCTS-AHD-Batch & 30745.4 (3.701\%) & 29691.2 (21.91\%) & 190046 (120.4\%) & 99940.9 (137.5\%) & 1142.85 (9.130\%) & 1022.69 (6.013\%) & 640.942 (8.311\%) & 1065.80 (10.03\%) \\
\hline
EoH-Split & 30856.0 (3.773\%) & 29301.7 (17.52\%) & 109201 (26.65\%) & 52918.3 (26.36\%) & 1137.64 (8.474\%) & 1023.26 (5.979\%) & 631.497 (6.570\%) & 1061.05 (9.467\%) \\
ReEvo-Split & 32248.0 (9.190\%) & 30328.2 (21.80\%) & 128112 (49.73\%) & 66069.2 (56.99\%) & 1157.77 (11.02\%) & 1028.40 (7.035\%) & 633.003 (7.203\%) & 1027.47 (6.414\%) \\
MCTS-AHD-Split & 32315.3 (9.706\%) & 31628.0 (38.50\%) & 197244 (125.1\%) & 99932.4 (137.5\%) & 1157.54 (10.73\%) & 1023.82 (6.331\%) & 647.133 (9.274\%) & 1061.06 (9.827\%) \\
\hline
\rowcolor{gray!15} \textbf{MuEvo} & \textbf{30045.4 (0.825\%)} & \textbf{26131.0 (3.233\%)} & \textbf{93933.2 (6.956\%)} & \textbf{47117.5 (17.46\%)} & \textbf{1081.74 (3.399\%)} & 993.109 (2.969\%) & \textbf{601.588 (2.100\%)} & \textbf{980.657 (1.052\%)} \\
\hline
\end{tabular}%
}
\centering
\caption{\textbf{Performance on the componentized ACO framework.} Values report mean cost and mean gap (\%). Cost and Gap are averaged over 3 independent runs. For all metrics, lower is better.}
\label{tab:aco_results}
\vspace{-4mm}
\end{table*}

\section{Experiments}
\label{sec:experiments}
We evaluate MuEvo on SHH and componentized ACO across four combinatorial optimization domains, comparing it with human-designed defaults and representative multi-component LLM-AHD extensions, and assessing cross-controller transfer, mechanism ablations, and robustness across LLM backbones.

\subsection{Experimental Setup}

\textbf{Frameworks and benchmarks.} We evaluate MuEvo on SHH and componentized ACO frameworks. The SHH framework uses a controller-mediated LLH pool and covers TSP, BPP, CVRP, and Flowshop, with 20/112, 25/88, 25/88, and 30/150 training/test instances from TSPLib, BPPLib, CVRPLib, and the VRF benchmark, respectively~\cite{TSPLIB:reinelt1991tsplib,BBPLIB:delorme2018bpplib,CVRPLib:uchoa2017new,VRFBench:vallada2015new}. The componentized ACO framework comprises functionally dependent components across the optimization workflow and uses the same TSP and CVRP partitions. Framework and component details are provided in Appendices A and B, while complete instance lists are provided in Appendix G.

\noindent \textbf{Baselines.} We compare MuEvo with the human-designed defaults, Default-SHH and Default ACO, and with representative learning-based methods where applicable: POMO~\cite{POMO:kwon2020policy} for TSP and CVRP, and DeepACO~\cite{ACO:ye2023deepaco} for componentized ACO. For multi-component LLM-AHD comparison, we further extend EoH, ReEvo, and MCTS-AHD. Each method performs fixed pre-search using its original search operators and feedback mechanism, and then optimizes the selected components in three modes: Single evolves one component with the others fixed; Split evolves multiple components independently and combines them afterward; and Batch jointly represents and updates the selected components. Details of these LLM-AHD baselines and their multi-component adaptations are provided in Appendix E.

\noindent \textbf{Training budget and MuEvo configuration.} All LLM-based methods use DeepSeek-V4-Flash through its official API with temperature 0.8 and the same generation limit. Following ReEvo's evaluation protocol~\cite{ReEvo:ye2024reevo}, each method receives 100 function evaluations (FEs) for the main evolutionary search. Before this search, MuEvo allocates 25 FEs to probe each valid component, while each LLM-AHD baseline performs its method-specific fixed pre-search with the same per-component budget. Thus, a framework with \(N\) valid components uses \(25N\) additional FEs in the preliminary stage, excluded from the 100-FE evolution budget. One FE evaluates one candidate on the full training set under a fixed solver-run budget per instance. Single-context methods spend this budget in one context, whereas MuEvo shares it across three collaboration contexts; hence, each regular or pair-evolution round consumes one FE. Lifecycle thresholds, context weights, pair scheduling, and adaptive-allocation parameters are reported in Appendix C.

\noindent \textbf{Evaluation protocol.} SHH candidates are evolved with ADAPHH as the controller. Each LLM-based method is run three times with different random seeds, and the results are averaged across these runs. For both SHH and ACO, each training instance is evaluated over 10 solver runs, while each test instance is evaluated over 10 independent runs with a 300-s time limit. We report mean solution cost, bin count, or makespan, together with the mean optimality gap to the best-known solution. Dataset-level results average instance gaps within each dataset, whereas aggregated analyses use instance-count-weighted means across datasets. Experiments are conducted on a server with two AMD EPYC 9754 processors and 1024 GB of RAM.

\subsection{Main Results}

Tables~\ref{tab:main_results} and~\ref{tab:aco_results} compare MuEvo with the human-designed defaults and three LLM-AHD baselines under SHH and componentized ACO, respectively. We examine whether MuEvo remains effective across the two frameworks despite their different component structures.

\subsubsection{SHH Results}

As shown in Table~\ref{tab:main_results}, MuEvo improves the original human-designed LLH ensemble on all 16 datasets and obtains the lowest or tied-lowest mean gap on 14. It performs best on three of the four TSP datasets, all four BPP datasets, three of the four CVRP datasets, and all four Flowshop datasets. Among the direct LLM-AHD extensions, Batch provides the strongest baseline on 11 datasets, whereas Split is competitive on several BPP and CVRP datasets but degrades markedly on medium- and large-scale TSP for ReEvo and MCTS-AHD. This instability is consistent with independently evolved LLHs becoming mismatched when recombined. MuEvo is more robust across domains, reducing the strongest baseline gap from 6.261\% to 5.237\% on TSP-L and from 3.506\% to 3.236\% on Flowshop VRF100.

\subsubsection{ACO Results}

Table~\ref{tab:aco_results} further evaluates MuEvo when the optimized components occupy different stages of the ACO workflow. MuEvo improves Default ACO on all eight datasets and achieves the lowest mean gap on seven, covering all four TSP datasets and three of the four CVRP datasets. The relative performance of the direct extensions again depends on the domain: Batch is strongest on the two smaller TSP sets, Split performs best on TSP-M and TSP-L, and Single is strongest on three CVRP sets. No fixed evolutionary mode therefore remains consistently effective as the component context changes. MuEvo instead yields the most consistent performance, including reductions from 26.65\% to 6.956\% on TSP-M and from 26.36\% to 17.46\% on TSP-L over the strongest LLM-AHD baselines. These results demonstrate that MuEvo remains effective across both controller-mediated LLH pools and functionally differentiated components with structured dependencies.

\begin{figure}[t]
    \centering
    \includegraphics[width=\columnwidth]{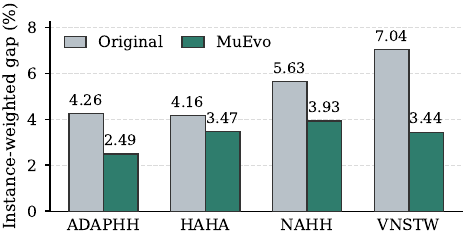}
    \caption{\textbf{Cross-controller generalization on TSP.} Instance-weighted mean gaps of the original and MuEvo-evolved LLH ensembles under four SHH controllers. Lower is better.}
    \label{fig:generalization_weighted}
    \vspace{-2mm}
\end{figure}

\subsection{Further Analysis}

\subsubsection{Cross-Controller Generalization}
\label{sec:rq2_generalization}

To examine whether the evolved LLH ensemble overfits the controller used during evolution, we train MuEvo with ADAPHH and transfer the resulting ensemble without modification to HAHA, NAHH, and VNSTW. These CHeSC 2011 controllers employ different heuristic-selection mechanisms~\citep{ADAPHH:misir2012intelligent,HAHA:lehrbaum2012new,NAHH:mascia2012non,VNSTW:hsiao2012vns}; their implementation details are provided in Appendix F.
As shown in Figure~\ref{fig:generalization_weighted}, the evolved ensemble reduces the instance-weighted TSP gap under all four controllers, with the largest improvement occurring under VNSTW, from 7.04\% to 3.44\%. Additional results on BPP, reported in Appendix C, show the same trend across all evaluated controllers. These results indicate that the ensembles evolved with ADAPHH transfer effectively to unseen controllers in both TSP and BPP without modification or retraining.

\subsubsection{Ablation Studies}

\begin{table}[t]
\resizebox{\columnwidth}{!}{%
\begin{tabular}{l|cc}
\hline
\multirow{2}{*}{\textbf{Method}} & \textbf{BPP-SHH} & \textbf{TSP-ACO} \\
\cline{2-3}
 & Gap (\%) $\downarrow$ & Gap (\%) $\downarrow$ \\
\hline
MuEvo (Full) & \textbf{0.7780 (+0.0000)} & \textbf{5.855 (+0.000)} \\
\hline
w/o Dynamic Component Management & 0.9516 (+0.1736) & 13.86 (+8.005) \\
w/o Multi-Ensemble Evaluation & 0.9268 (+0.1488) & 12.83 (+6.975) \\
w/o Cross-Component Information Sharing & 1.1574 (+0.3794) & 9.999 (+4.144) \\
w/o Relation-Guided Pair Evolution & 1.1669 (+0.3889) & 12.11 (+6.255) \\
w/o Adaptive Budget Allocation & 0.8598 (+0.0818) & 10.75 (+4.895) \\
\hline
\end{tabular}%
}
\centering
\caption{Ablation study reporting instance-weighted gaps and absolute increases over MuEvo (in parentheses). Lower is better.}
\label{tab:ablation_gap_diffs}
\end{table}



Table~\ref{tab:ablation_gap_diffs} evaluates five key mechanisms of MuEvo on BPP-SHH and TSP-ACO. Each variant removes one mechanism while keeping the probing results, evaluation protocol, and search budget unchanged. On TSP-ACO, removing Dynamic Component Management causes the largest degradation, increasing the weighted gap from 5.855\% to 13.86\%. Removing Multi-Ensemble Evaluation, Cross-Component Information Sharing, or Relation-Guided Pair Evolution also substantially worsens performance, yielding gaps of 12.83\%, 9.999\%, and 12.11\%, respectively. The trend is consistent on BPP-SHH, where every ablation increases the gap over MuEvo. Cross-Component Information Sharing and Relation-Guided Pair Evolution have the largest effects, while replacing Adaptive Budget Allocation with random selection increases the gap from 0.7780\% to 0.8598\%. These results demonstrate that both Dynamic Component Management and the co-evolution mechanisms are important to MuEvo.

\subsubsection{MuEvo with Different LLMs}
To examine the sensitivity of MuEvo to the underlying LLM, we evaluate three LLM backbones on BPP-SHH and TSP-ACO under the same search and evaluation settings. All models are accessed through their official APIs. As shown in Table~\ref{tab:llm_study}, all three backbones substantially improve over the original heuristics in both settings. The best-performing model varies by framework: DeepSeek-V4-Flash achieves the lowest BPP-SHH gap of 0.778\%, whereas Kimi-K2.6 achieves the lowest TSP-ACO gap of 3.588\%. These results indicate that MuEvo remains effective across different LLM backbones.
\begin{table}[t]
\resizebox{\columnwidth}{!}{%
\begin{tabular}{l|cc}
\hline
\multirow{2}{*}{\textbf{LLM}} & \textbf{BPP-SHH} & \textbf{TSP-ACO} \\
\cline{2-3}
 & Bins (Gap) $\downarrow$ & Cost (Gap) $\downarrow$ \\
\hline
Ori-Heuristic & 88.1830 (3.113\%) & 103922 (93.28\%) \\
\hline
DeepSeek-V4-Flash & \textbf{86.1887 (0.778\%)} & 49264.5 (5.855\%) \\
Kimi-K2.6 & 86.4614 (1.192\%) & \textbf{48506.6 (3.588\%)} \\
Qwen3.7-Plus-2026-05-26 & 86.2648 (0.981\%) & 47087.0 (5.728\%) \\
\hline
\end{tabular}%
}
\centering
\caption{MuEvo with different LLM backbones. Values report instance-weighted cost and gap (\%). Lower is better.}
\label{tab:llm_study}
\vspace{-4mm}
\end{table}

%% file: Sections/5_Conclusion.tex
\section{Conclusion}
\label{sec:conclusion}

In this paper, we study multi-heuristic LLM-AHD, where fixed pre-search may miss components with late potential and independent evolution neglects their dependencies. We propose MuEvo, which combines Dynamic Component Management to revise component priorities with LLM-Driven Co-Evolution to coordinate component populations through ensemble feedback. Experiments on SHH and componentized ACO across four combinatorial optimization domains show that consistently outperforms strong baselines. Cross-controller transfer and ablation studies further support its generality and core mechanisms.


%% file: Sections/6_Appendix.tex

\appendix


\input{Sections/AppendixDetail/SHHDetail.tex}
\input{Sections/AppendixDetail/ACODetail.tex}

\input{Sections/AppendixDetail/ExperimentDetail.tex}

\input{Sections/AppendixDetail/PromptDetail.tex}

\section{LLM-Based Automatic Heuristic Design Baselines}
\label{app:baseline_methods}

We compare MuEvo with three representative LLM-based automatic heuristic design methods: ReEvo, EoH, and MCTS-AHD. This section summarizes their original search mechanisms and then describes how they are instantiated in the Single, Split, and Batch modes used in our experiments.

\subsection{ReEvo: Reflective Evolution Framework}
\label{app:reevo}

This section provides a detailed overview of the ReEvo framework~\citep{ReEvo:ye2024reevo}, which pioneered the concept of Language Hyper-Heuristics (LHH) for automatic heuristic design. Our MuEvo framework builds upon ReEvo's foundational ideas while extending them to the multi-heuristic co-evolution setting.

\subsubsection{Language Hyper-Heuristics}
ReEvo introduces the concept of \textit{Language Hyper-Heuristics} (LHH), an emerging variant of traditional Hyper-Heuristics that leverages Large Language Models for heuristic generation. Unlike classic HHs that are limited by heuristic spaces predefined by human experts, LHH features minimal human intervention and open-ended heuristic spaces.

Formally, for a combinatorial optimization problem with solution space $\mathcal{S}$ and objective function $f: \mathcal{S} \rightarrow \mathbb{R}$, a Hyper-Heuristic searches for the optimal heuristic $h^*$ in a heuristic space $\mathcal{H}$ such that a meta-objective function $F: \mathcal{H} \rightarrow \mathbb{R}$ is minimized:
\begin{equation}
    h^* = \arg\min_{h \in \mathcal{H}} F(h)
\end{equation}
In LHH, heuristics in $\mathcal{H}$ are generated by LLMs rather than being predefined, enabling exploration of an open-ended heuristic space.

\subsubsection{ReEvo Methodology}
ReEvo couples evolutionary search with LLM self-reflections to efficiently explore heuristic spaces. The framework employs two LLM roles: a \textbf{generator LLM} for generating heuristic candidates, and a \textbf{reflector LLM} for providing guidance through reflections.

\paragraph{Individual Encoding.}
ReEvo performs evolutionary search over executable heuristic code. Each individual is a code snippet representing a heuristic function and is constrained primarily by the required function signature rather than a predefined program encoding.

\paragraph{Population Initialization.}
The heuristic population is initialized by prompting the generator LLM with a task specification containing: (1) COP descriptions (if available), (2) heuristic designation, and (3) heuristic functionality. Optionally, seed heuristics can provide in-context examples to encourage valid generation.

\paragraph{Iterative Evolution.}
Each ReEvo iteration contains five sequential steps:
\begin{enumerate}
    \item \textbf{Selection:} Parent pairs are selected from successfully executed heuristics at random, while avoiding pairing heuristics with identical meta-objective values.
    \item \textbf{Short-term Reflection:} The reflector LLM analyzes the relative performance of each parent pair and provides hints for improved design. This is analogous to interpreting genetic cues and providing ``verbal gradients'' within the search space.
    \item \textbf{Crossover:} The generator LLM produces an offspring heuristic given the parent pair, their relative performance, short-term reflections, and generation instructions.
    \item \textbf{Long-term Reflection:} The reflector LLM summarizes accumulated short-term reflections into long-term insights, distilling expertise for improved heuristic design.
    \item \textbf{Elitist Mutation:} Based on long-term reflections, the generator LLM samples multiple heuristics to improve the current best one.
\end{enumerate}

The dual-level reflection mechanism—short-term for pairwise comparison and long-term for accumulated expertise—enables ReEvo to outperform prior state-of-the-art LHH methods with better sample efficiency.

\begin{algorithm}[t]
\caption{ReEvo Workflow}
\label{alg:reevo}
\begin{algorithmic}[1]
\STATE \textbf{Input:} Task Specification $\mathcal{T}$, Optional Seed Heuristics $\mathcal{H}_{seed}$
\STATE \textbf{Initialize:} Population $\mathcal{P} \leftarrow \text{LLM}(\mathcal{T}, \mathcal{H}_{seed})$
\STATE Long-term reflections $\mathcal{R}_{long} \leftarrow \emptyset$
\WHILE{not converged}
    \STATE \textbf{Evaluate:} Compute meta-objective $F(h)$ for all $h \in \mathcal{P}$
    \STATE \textbf{Selection:} Select parent pairs $(h_1, h_2)$ from $\mathcal{P}$
    \FOR{each parent pair $(h_1, h_2)$}
        \STATE \textbf{Short-term Reflection:} $r_{short} \leftarrow \text{Reflector}(h_1, h_2, F(h_1), F(h_2))$
        \STATE \textbf{Crossover:} $h_{new} \leftarrow \text{Generator}(h_1, h_2, r_{short})$
    \ENDFOR
    \STATE \textbf{Long-term Reflection:} $\mathcal{R}_{long} \leftarrow \text{Reflector}(\mathcal{R}_{long}, \{r_{short}\})$
    \STATE \textbf{Elitist Mutation:} $h_{mutated} \leftarrow \text{Generator}(h^*, \mathcal{R}_{long})$ where $h^* = \arg\min F$
    \STATE \textbf{Update:} $\mathcal{P} \leftarrow \text{Select}(\mathcal{P} \cup \{h_{new}\} \cup \{h_{mutated}\})$
\ENDWHILE
\STATE \textbf{Return:} Best Heuristic $h^* \in \mathcal{P}$
\end{algorithmic}
\end{algorithm}

\subsubsection{Relationship to MuEvo}
\label{subsec:reevo_comparison}

MuEvo retains ReEvo's reflection-guided within-population evolution while
addressing the additional problem of coordinating multiple components in a
complete solver. The key distinction is the shift from
\textit{single-heuristic evolution} to \textit{multi-component
co-evolution}:

\begin{itemize}
    \item \textbf{ReEvo} focuses on evolving a single heuristic function for a specific component of the problem solver (e.g., a penalty heuristic for GLS, or a heuristic measure for ACO).
    \item \textbf{MuEvo}\ maintains one population for each component
    and evaluates candidate replacements in the complete SHH or
    componentized ACO solver.
\end{itemize}

This difference motivates two groups of extensions:
\begin{enumerate}
    \item \textbf{Dynamic Component Management} uses short-budget
    probing and a reversible lifecycle so that an initially weak
    component is not permanently removed.
    \item \textbf{LLM-Driven Co-Evolution} adds Multi-Ensemble
    Evaluation, cross-component relation context, relation-guided pair
    evolution, and adaptive budget allocation. These mechanisms use
    ensemble-level replacement evidence rather than treating component
    populations as independent searches.
\end{enumerate}

The generator and reflector prompts remain ReEvo-style, but component
eligibility, collaboration context, joint updates, and allocation are
controlled by MuEvo. Their exact configuration is reported in Appendix~C
and their prompts in Appendix~D.

\input{Sections/AppendixDetail/EOHDetail.tex}

\subsection{MCTS-AHD: Monte Carlo Tree Search for Heuristic Design}
\label{app:mcts_ahd}

The third baseline is MCTS-AHD~\citep{MCTS-AHD:zheng2025monte}, a recent LLM-based Automatic Heuristic Design method that addresses the local optima problem of population-based approaches through Monte Carlo Tree Search.

\subsubsection{Motivation and Key Insight}
Population-based LLM-AHD methods (such as ReEvo and EoH) maintain a fixed-size population of top-performing heuristics, discarding inferior ones in each generation. However, temporarily underperforming heuristics may still have potential for significant improvement after further LLM-based refinement. MCTS-AHD addresses this by preserving \emph{all} generated heuristics in a tree structure, enabling ``worse-before-better'' evolutionary paths.

\subsubsection{MCTS-AHD Methodology}

\paragraph{Tree Structure.}
MCTS-AHD organizes all LLM-generated heuristics in a tree where:
\begin{itemize}
    \item The root node $n_r$ is a virtual node without representing any heuristic
    \item Each non-root node represents an executable heuristic function $h$ along with its linguistic description
    \item Parent-child relationships encode evolutionary derivation (child generated from parent)
\end{itemize}

\paragraph{LLM-Based Actions.}
MCTS-AHD defines six types of LLM-based actions for node expansion:
\begin{itemize}
    \item \textbf{Initialization ($i_1$):} Generate a heuristic function from scratch
    \item \textbf{Mutation ($m_1$):} Introduce new mechanisms/formulas to existing function
    \item \textbf{Mutation ($m_2$):} Modify parameter settings of existing function
    \item \textbf{Crossover ($e_1$):} Generate divergent heuristic from multiple existing ones
    \item \textbf{Crossover ($e_2$):} Improve parent heuristic by learning from reference function
    \item \textbf{Tree-path Reasoning ($s_1$):} Analyze all heuristics on path from leaf to root and generate enhanced version
\end{itemize}

\paragraph{UCT Selection with Exploration-Decay.}
Node selection uses the UCT formula with normalized quality values:
\begin{equation}
    \text{UCT}(c) = \frac{Q(c) - q_{min}}{q_{max} - q_{min}} + \lambda \cdot \sqrt{\frac{\ln(N(n_c) + 1)}{N(c)}}
\end{equation}
where $Q(c)$ is the quality value (heuristic performance), $N(c)$ is the visit count, and $\lambda$ is the exploration factor. MCTS-AHD applies \textbf{exploration-decay} by linearly decreasing $\lambda$ as iterations progress:
\begin{equation}
    \lambda = \lambda_0 \cdot \frac{T - t}{T}
\end{equation}
This encourages exploration in early stages and exploitation in later stages.

\paragraph{Progressive Widening.}
To enable crossover between heuristics generated at different times, MCTS-AHD applies progressive widening, adding new child nodes when:
\begin{equation}
    \lfloor N(n)^\alpha \rfloor \geq |\text{Children}(n)|
\end{equation}
with $\alpha = 0.5$, allowing re-exploration of non-leaf nodes as the elite set evolves.

\begin{algorithm}[t]
\caption{MCTS-AHD Workflow}
\label{alg:mcts_ahd}
\begin{algorithmic}[1]
\STATE \textbf{Input:} Task $P$, Framework, Max evaluations $T$
\STATE \textbf{Initialize:} Tree with root $n_r$ and $N_I$ initial nodes via action $i_1$
\STATE Elite set $E \leftarrow \text{Top-10 performing heuristics}$
\WHILE{total evaluations $< T$}
    \STATE \textbf{Selection:} Traverse from $n_r$ to leaf $n_l$ by max UCT
    \IF{Progressive widening condition met}
        \STATE Expand $n_r$ with $e_1$ or other nodes with $e_2$
    \ENDIF
    \STATE \textbf{Expansion:} Generate $2k+2$ children via $m_1, m_2, e_2, s_1$
    \STATE \textbf{Simulation:} Evaluate each child heuristic $h$: $Q(c) \leftarrow g(h)$
    \STATE \textbf{Backpropagation:} Update $Q, N$ on path to root
    \STATE Update elite set $E$, $q_{max}$, $q_{min}$
\ENDWHILE
\STATE \textbf{Return:} Best heuristic $h^* = \arg\max_{h} g(h)$
\end{algorithmic}
\end{algorithm}

\subsection{Multi-Heuristic Baseline Adaptations}

To evaluate MuEvo's multi-heuristic co-evolution capability, we instantiate
the baseline methods in three modes. All three modes use the same component
ranking produced by their fixed pre-search. \textit{Single} selects only the
Top-1 component and evolves it while keeping the remaining ensemble fixed.
\textit{Split} and \textit{Batch} both select the same Top-3 components. Split
runs an independent single-component search for each selected target and
combines their best outputs only after all three searches finish, whereas
Batch treats the Top-3 set as a single optimization unit and prompts the LLM
to generate or refine all three components simultaneously.

\begin{table}[t]
\centering
\caption{Target selection and main-search budgets of the multi-heuristic
baseline adaptations. The preliminary pre-search budget is reported
separately in Appendix~C.}
\label{tab:baseline_budget_allocation}
\scriptsize
\renewcommand{\arraystretch}{1.12}
\begin{tabular}{@{}p{0.13\columnwidth}p{0.15\columnwidth}p{0.42\columnwidth}p{0.13\columnwidth}@{}}
\toprule
\textbf{Mode} & \textbf{Targets} & \textbf{Searches} & \textbf{Total FEs} \\
\midrule
Single & Top-1 & One 100-FE search & 100 \\
Batch & Top-3 & One joint 100-FE search & 100 \\
Split & Top-3 & Three independent 100-FE searches & 300 \\
\bottomrule
\end{tabular}
\end{table}

For Single and Batch, the 100-FE budget applies to the complete main search,
not separately to each selected component. In Batch, one FE evaluates one
joint Top-3 candidate. Split does not divide 100 FEs among its three targets:
each independent component search receives the full 100 FEs, for 300 main
search FEs in total. Consequently, no remainder-allocation rule is needed.
This deliberately generous allocation favors Split and tests whether
additional independent search can compensate for the absence of explicit
co-evolution; its comparatively weak results therefore cannot be attributed
to a smaller evaluation budget.

Across the Batch adaptations, a response is valid only when all three
required named functions can be extracted and pass the corresponding code
checks. An incomplete or invalid multi-function response is discarded as one
failed Batch proposal and is never partially merged with the current
ensemble. Because it is not sent to the solver evaluator, it consumes no FE;
the baseline proceeds to its next scheduled generation attempt.

\subsubsection{MCTS-AHD-Batch}
MCTS-AHD-Batch extends the standard MCTS-AHD framework to accommodate the simultaneous evolution of a heuristic ensemble. Unlike the standard version that searches for a single optimal heuristic, MCTS-AHD-Batch treats the entire ensemble $\{h_1, \ldots, h_n\}$ as a single optimization unit within the MCTS tree.

\paragraph{Key Design.}
\begin{itemize}
    \item \textbf{Ensemble-Level Search Space:} Each node in the MCTS tree represents a candidate ensemble. The expansion and simulation operations are modified to generate and evaluate all heuristics in the ensemble simultaneously.
    \item \textbf{Joint Generation:} During the expansion phase, the LLM is prompted to generate code for all target heuristics in the ensemble in a single pass, ensuring the LLM can consider the potential interactions between different operators.
    \item \textbf{Ensemble Evaluation:} The reward signals (simulation results) are derived from the performance of the complete ensemble when integrated into the hyper-heuristic framework, allowing the search to optimize for synergistic effects.
\end{itemize}

\begin{algorithm}[t]
\caption{MCTS-AHD-Batch Workflow}
\label{alg:mcts_ahd_batch}
\begin{algorithmic}[1]
\STATE \textbf{Input:} Target LLH IDs $\{h_1, \ldots, h_n\}$, Max evaluations $T$
\STATE \textbf{Initialize:} Tree with root $n_r$ representing the initial ensemble
\WHILE{total evaluations $< T$}
    \STATE \textbf{Selection:} Traverse from $n_r$ to a leaf node $n_l$ using the UCT criterion
    \STATE \textbf{Expansion:} Prompt LLM to generate a new ensemble $\mathcal{E}_{new} = \{h_1, \ldots, h_n\}$ in a single pass
    \STATE \textbf{Simulation:} Evaluate $\mathcal{E}_{new}$ in the hyper-heuristic framework to obtain fitness $Q(\mathcal{E}_{new})$
    \STATE \textbf{Backpropagation:} Update $Q$ and $N$ values for all nodes on the path from $n_l$ to $n_r$
\ENDWHILE
\STATE \textbf{Return:} Best ensemble $\mathcal{E}^* = \arg\max_{\mathcal{E}} Q(\mathcal{E})$
\end{algorithmic}
\end{algorithm}

\subsubsection{ReEvo-Batch}
ReEvo-Batch extends ReEvo to evolve an ensemble of multiple LLHs simultaneously within a single evolutionary process.

\paragraph{Key Design.}
\begin{itemize}
    \item \textbf{Multi-LLH Individual:} Unlike standard ReEvo where each individual is a single heuristic, ReEvo-Batch represents each individual as an \emph{ensemble} of multiple LLHs: $\text{Individual} = \{h_1, h_2, \ldots, h_n\}$.
    \item \textbf{Ensemble Evaluation:} The fitness of each individual is evaluated by injecting all its LLHs into the hyper-heuristic framework and measuring ensemble-level performance.
    \item \textbf{Dual-Level Reflection:} Maintains both short-term reflection (before crossover) and long-term reflection (accumulated insights for mutation), applied to the entire ensemble.
    \item \textbf{Single-Pass Generation:} During evolutionary operators (crossover and mutation), the LLM processes and generates code for all heuristics in the ensemble within a single prompt-response cycle, rather than updating them one by one.
    \item \textbf{Parallel Evaluation:} Uses ProcessPoolExecutor for efficient parallel fitness evaluation of ensemble individuals.
\end{itemize}

\begin{algorithm}[t]
\caption{ReEvo-Batch Workflow}
\label{alg:reevo_batch}
\begin{algorithmic}[1]
\STATE \textbf{Input:} Target LLH IDs $\{h_1, \ldots, h_n\}$, Population size $M$
\STATE \textbf{Initialize:} Population $\mathcal{P} \leftarrow \{(\{h_1, \ldots, h_n\})_1, \ldots, (\{h_1, \ldots, h_n\})_M\}$
\STATE Long-term reflection $\mathcal{R}_{long} \leftarrow \emptyset$
\WHILE{not converged}
    \STATE \textbf{Evaluate:} Compute ensemble fitness for all individuals (parallel)
    \STATE \textbf{Selection:} Select parent pairs from $\mathcal{P}$
    \FOR{each parent pair $(P_1, P_2)$}
        \STATE \textbf{Short-term Reflection:} $r_{short} \leftarrow \text{Reflector}(P_1, P_2)$
        \STATE \textbf{Crossover:} $P_{new} \leftarrow \text{Generator}(P_1, P_2, r_{short})$
    \ENDFOR
    \STATE \textbf{Long-term Reflection:} $\mathcal{R}_{long} \leftarrow \text{Reflector}(\mathcal{R}_{long}, \{r_{short}\})$
    \STATE \textbf{Elitist Mutation:} $P_{mutated} \leftarrow \text{Generator}(P^*, \mathcal{R}_{long})$
    \STATE \textbf{Update:} $\mathcal{P} \leftarrow \text{Select}(\mathcal{P} \cup \{P_{new}\} \cup \{P_{mutated}\})$
\ENDWHILE
\STATE \textbf{Return:} Best ensemble $P^* \in \mathcal{P}$
\end{algorithmic}
\end{algorithm}

\subsubsection{EoH-Single, EoH-Split, and EoH-Batch}
EoH-Single follows the standard EoH setting and evolves one target LLH with one population and one evaluator. EoH-Split launches an independent EoH-Single search for each selected LLH and combines the best resulting functions after all runs terminate; populations, prompts, and fitness feedback are not shared across targets. EoH-Batch is a portfolio-level extension in which each individual contains all target LLHs, its natural-language description summarizes the complete portfolio, and its fitness is obtained by evaluating the complete ensemble in the target framework. EoH-Batch retains the E1, E2, M1, and M2 operator semantics, rank-based parent selection, and elitist population survival of EoH, but changes the individual representation from one function to a set of functions generated in a single LLM response.

\subsection{Comparison with MuEvo}
The batch baselines generate ensembles of LLHs in a single-pass manner, but differ fundamentally from MuEvo in how they handle the expanded search space:

\begin{itemize}
    \item \textbf{Batch Baselines (EoH-Batch, MCTS-AHD-Batch, and ReEvo-Batch):} These methods treat the ensemble as a single atomic unit. All target heuristics are generated or updated simultaneously in one LLM response. While this exposes the complete ensemble to the model, it requires each evolutionary step to search over the joint code space of all target heuristics.
    \item \textbf{MuEvo:} Employs LLM-Driven Co-Evolution, maintaining separate component populations while coordinating them through Cross-Component Information Sharing, Relation-Guided Pair Evolution, Multi-Ensemble Evaluation, and Adaptive Budget Allocation.
\end{itemize}

\section{Details of Selection Hyper-Heuristics (SHH)}
\label{app:shh_details}

This section details the four Selection Hyper-Heuristic (SHH) algorithms used
in our cross-controller generalization experiments. These algorithms represent
diverse search paradigms: adaptive dynamic heuristic sets (ADAPHH), online
learning automata (HAHA), offline racing (NAHH), and variable neighborhood
search (VNSTW).

We use Python ports of the released CHeSC implementations and retain their
default controller constants. For every portfolio comparison, the original
and evolved LLH ensembles use the same instance, solver seed, and time limit;
only the LLH implementations are exchanged. The transfer experiments do not
retrain or retune any controller. Algorithm-specific fixed schedules are
shown below, including HAHA's seven parallel offspring and NAHH's phase
fractions.

\subsection{ADAPHH (Adaptive Dynamic Heuristic Set)}
ADAPHH~\citep{ADAPHH:misir2012intelligent} introduces an adaptive mechanism to maintain a dynamic subset of high-performing heuristics. It employs Relay Hybridization to discover effective heuristic pairs and learning automata for selection.

\begin{algorithm}[t]
\caption{ADAPHH Workflow}
\label{alg:adaphh}
\begin{algorithmic}[1]
\STATE \textbf{Initialize:} Phase weights, Quality Index (QI), Tabu list $\mathcal{T} \leftarrow \emptyset$
\WHILE{not converged}
    \STATE \textbf{Learning Phase:} Update QI for single heuristics
    \STATE \textbf{Subset Selection:} Exclude heuristics with poor QI
    \STATE \textbf{Relay Hybridization:}
    \STATE Select heuristic $h_i$ based on performance memory
    \STATE Select successor $h_j$ based on transition probability $P(h_j|h_i)$
    \STATE Apply sequence $(h_i, h_j)$ to solution $S$
    \STATE \textbf{Acceptance:} Accept if improving or via Simulated Annealing
    \STATE Update performance metrics and transition probabilities
    \STATE \textbf{Tabu Mechanism:} Move stagnating heuristics to $\mathcal{T}$
\ENDWHILE
\end{algorithmic}
\end{algorithm}

\subsection{HAHA (History-based Adaptive Heuristic Allocation)}
HAHA~\citep{HAHA:lehrbaum2012new} is an online-learning selection hyper-heuristic that alternates
between a \emph{serial search} phase on a single working solution and a \emph{parallel search} phase
over several mutated candidate solutions. It maintains quality scores for local-search heuristics,
applies them in decreasing-quality order, periodically updates these qualities, and uses a solution
ringbuffer to avoid cycling. 

\begin{algorithm}[t]
\caption{HAHA Workflow (paper-faithful, simplified)}
\label{alg:haha}
\begin{algorithmic}[1]
\STATE \textbf{Init:} Create ringbuffer $b$; initialise heuristic qualities $q$; sort local-search heuristics $LS$ by decreasing $q$.
\STATE $s_{\text{work}} \leftarrow \textsc{InitialiseLSQualities}()$, \ \ $s_{\text{best}} \leftarrow s_{\text{work}}$.
\WHILE{time not expired}
    \STATE $b.\text{add}(s_{\text{work}})$
    \STATE \textbf{Serial Search:} $s_{\text{work}} \leftarrow \textsc{SerialSearch}(s_{\text{work}}, LS, q, b)$
    \STATE \textbf{Generate Mutations:} create $m$ mutated offsprings $\{s_1,\ldots,s_m\}$ from $s_{\text{work}}$ (paper uses $m=7$)  \COMMENT{roulette-wheel over mutation qualities} 
    \STATE \textbf{Parallel Search:} apply scheduled local search to each $s_i$; 
           \IF{a global improvement ($s_i$ better than $s_{\text{best}}$) is found}
               \STATE $s_{\text{best}} \leftarrow s_i$, \ $s_{\text{work}} \leftarrow s_i$; \textbf{abort} parallel search and continue
           \ENDIF
    \STATE \textbf{Working Solution Selection:} if no global improvement, set $s_{\text{work}} \leftarrow \textsc{SelectSolution}(\{s_1,\ldots,s_m\})$
\ENDWHILE
\STATE \textbf{return} $s_{\text{best}}$
\end{algorithmic}
\end{algorithm}

3

\subsection{NAHH (Non-Adaptive Hyper-Heuristic)}
NAHH~\citep{NAHH:mascia2012non} leverages an offline configuration or "racing" phase to select a static subset of heuristics or a fixed sequence that performs best on training instances, then applies it during the search.

\begin{algorithm}[t]
\caption{NAHH Workflow}
\label{alg:nahh}
\begin{algorithmic}[1]
\STATE \textbf{Input:} target instance $I$, low-level heuristics $\mathcal{H}$, total time budget $T$
\STATE \textbf{Offline (before deployment):} prepare a small pool of fixed-parameter schemata $\mathcal{S}=\{s_1,\dots,s_K\}$ tuned on sample domains
\STATE \textbf{Phase 1 (LLH analysis, $\le 0.075T$):}
\FOR{each $h \in \mathcal{H}$}
  \STATE Run $h$ on several random starts; record median runtime $t_h$ and median quality $q_h$
\ENDFOR
\STATE Compute non-dominated set $\mathcal{H}' \subseteq \mathcal{H}$ using dominance on $(t_h,q_h)$; keep at least the two best-quality heuristics
\STATE Obtain the best solution found so far $S_{\text{best}}$ from this phase

\STATE \textbf{Phase 2 (Schema selection by online race, up to $0.25T$):}
\STATE Active set $\mathcal{A}\leftarrow \mathcal{S}$; initialise all schemata from $S_{\text{best}}$
\WHILE{$|\mathcal{A}|>1$ \AND time used in Phase 2 $< 0.25T$}
  \FOR{each schema $s \in \mathcal{A}$ (interleaved)}
     \STATE Run $s$ for a short slice of time (e.g., $0.025\cdot (T - T_1)$) using heuristics in $\mathcal{H}'$
     \STATE Update its best-achieved quality
  \ENDFOR
  \STATE Eliminate the worst-performing schema from $\mathcal{A}$
\ENDWHILE
\STATE Let $s^\star$ be the remaining (or best-so-far) schema

\STATE \textbf{Phase 3 (Run):}
\STATE Execute $s^\star$ for the remaining time budget using $\mathcal{H}'$
\STATE \textbf{Return:} best solution found
\end{algorithmic}
\end{algorithm}

\subsection{VNSTW (Variable Neighborhood Search with Time Windows)}
VNSTW~\citep{VNSTW:hsiao2012vns} adapts the classic VNS by dynamic management of neighborhood ordering and shaking intensities.

\begin{algorithm}[t]
\caption{VNSTW Workflow}
\label{alg:vnstw}
\begin{algorithmic}[1]
\STATE \textbf{Initialize:} Neighborhoods $N_1, \dots, N_k$, Ordering $\pi$
\STATE Select random initial solution $S$
\WHILE{not converged}
    \STATE $k \leftarrow 1$
    \WHILE{$k \leq k_{max}$}
        \STATE \textbf{Shaking:} Generate $S'$ in neighborhood $N_{\pi(k)}(S)$
        \STATE \textbf{Local Search:} $S'' \leftarrow \text{LocalSearch}(S')$
        \STATE \textbf{Move:}
        \IF{$f(S'') < f(S)$}
            \STATE $S \leftarrow S''$
            \STATE $k \leftarrow 1$ (Reset neighborhood)
        \ELSE
            \STATE $k \leftarrow k + 1$ (Next neighborhood)
        \ENDIF
        \STATE \textbf{Adaptation:} Update success counts for $N_{\pi(k)}$
        \STATE Periodically reorder $\pi$ based on success rates
    \ENDWHILE
\ENDWHILE
\end{algorithmic}
\end{algorithm}

\section{Benchmark Datasets}
\label{app:datasets}

This appendix details the dataset configurations utilized in our multi-heuristic experiments, covering TSP, Bin Packing, Flowshop, and CVRP domains.

\subsection{TSP}
We use TSPLib benchmark instances~\citep{TSPLIB:reinelt1991tsplib}
organized into four test sets categorized by instance size. All instances use
Euclidean distance and contain optimal or best-known solutions for performance
evaluation.

\subsubsection{Test Sets}
We evaluate on four progressively larger test sets to assess scalability and generalization: TSP-SS (directory \texttt{FtsplibSS}; Extra Small), TSP-S (directory \texttt{FtsplibS}; Small), TSP-M (directory \texttt{FtsplibM}; Medium), and TSP-L (directory \texttt{FtsplibL}; Large). Table~\ref{tab:tsp-test-all} consolidates all 112 test instances.

\begin{table}[t]
  \centering
  \caption{The 112 test instances in TSP-SS, TSP-S, TSP-M, and TSP-L.}
  \label{tab:tsp-test-all}
  \scriptsize
  \resizebox{\linewidth}{!}{
  \begin{tabular}{@{}lllll@{}}
    \toprule
    \multicolumn{5}{c}{\textbf{TSP-SS (\texttt{FtsplibSS}; 47 instances; 100--500 cities)}} \\
    \midrule
    a280 (280) & bcl380 (380) & bier127 (127) & ch130 (130) & ch150 (150) \\
    d198 (198) & d493 (493) & eil101 (101) & fl417 (417) & gil262 (262) \\
    kroA100 (100) & kroA150 (150) & kroA200 (200) & kroB100 (100) & kroB150 (150) \\
    kroB200 (200) & kroC100 (100) & kroD100 (100) & kroE100 (100) & lin105 (105) \\
    lin318 (318) & linhp318 (318) & pbk411 (411) & pbl395 (395) & pbm436 (436) \\
    pbn423 (423) & pcb442 (442) & pka379 (379) & pma343 (343) & pr107 (107) \\
    pr124 (124) & pr136 (136) & pr144 (144) & pr152 (152) & pr226 (226) \\
    pr264 (264) & pr299 (299) & pr439 (439) & qa194 (194) & rat195 (195) \\
    rd100 (100) & rd400 (400) & ts225 (225) & tsp225 (225) & u159 (159) \\
    xqf131 (131) & xqg237 (237) &  &  &  \\
    \midrule
    \multicolumn{5}{c}{\textbf{TSP-S (\texttt{FtsplibS}; 15 instances; 500--1000 cities)}} \\
    \midrule
    d657 (657) & dkg813 (813) & lim963 (963) & lu980 (980) & p654 (654) \\
    pbd984 (984) & rat575 (575) & rat783 (783) & rbu737 (737) & rbx711 (711) \\
    u574 (574) & u724 (724) & uy734 (734) & xql662 (662) & zi929 (929) \\
    \midrule
    \multicolumn{5}{c}{\textbf{TSP-M (\texttt{FtsplibM}; 29 instances; 1000--2000 cities)}} \\
    \midrule
    d1291 (1291) & d1655 (1655) & dca1389 (1389) & dcc1911 (1911) & dja1436 (1436) \\
    djc1785 (1785) & dka1376 (1376) & dkd1973 (1973) & fl1400 (1400) & fl1577 (1577) \\
    fnb1615 (1615) & fra1488 (1488) & icw1483 (1483) & mu1979 (1979) & nrw1379 (1379) \\
    pcb1173 (1173) & pr1002 (1002) & rbv1583 (1583) & rby1599 (1599) & rl1304 (1304) \\
    rl1323 (1323) & rl1889 (1889) & rw1621 (1621) & u1060 (1060) & u1432 (1432) \\
    u1817 (1817) & vm1084 (1084) & vm1748 (1748) & xit1083 (1083) &  \\
    \midrule
    \multicolumn{5}{c}{\textbf{TSP-L (\texttt{FtsplibL}; 21 instances; 2000--5000 cities)}} \\
    \midrule
    bch2762 (2762) & bck2217 (2217) & beg3293 (3293) & bva2144 (2144) & d2103 (2103) \\
    dbj2924 (2924) & dcb2086 (2086) & dea2382 (2382) & djb2036 (2036) & irw2802 (2802) \\
    ley2323 (2323) & lsm2854 (2854) & mlt2597 (2597) & pds2566 (2566) & pr2392 (2392) \\
    rbw2481 (2481) & u2152 (2152) & u2319 (2319) & xpr2308 (2308) & xqc2175 (2175) \\
    xva2993 (2993) &  &  &  &  \\
    \bottomrule
  \end{tabular}}
\end{table}

\subsection{Bin Packing}
We utilize a diverse collection of Bin Packing Problem (BPP) instances from well-known benchmarks, including datasets from Scholl~\cite{Scholl:scholl1997bison}, Schwerin~\cite{BPPS:schwerin1997bin}, and Falkenauer~\cite{BPPF:falkenauer1996hybrid}. All instances are 1D Bin Packing problems where the goal is to minimize the number of bins used.

\subsubsection{Test Sets}
We evaluate on four distinct test sets covering varied problem characteristics: Scholl Set 3 (Hard), Schwerin Set 3, Falkenauer U (Uniform), and Falkenauer T (Triplets). Table~\ref{tab:bp-test-all} consolidates all instances.

\begin{table}[t]
  \centering
  \caption{Combined Bin Packing Test Instances (Scholl, Schwerin, Falkenauer U/T).}
  \label{tab:bp-test-all}
  \scriptsize
  \resizebox{\linewidth}{!}{
  \begin{tabular}{@{}ll@{}}
    \toprule
    \textbf{Instance} (\textit{Detail: C=Cap, N=Items, [Range]}) & \textbf{Instance} (\textit{Detail: C=Cap, N=Items, [Range]}) \\
    \midrule
    \multicolumn{2}{c}{\textbf{Scholl Set 3 (Hard) -- 10 instances}} \\
    \midrule
    HARD0 (C=100000, N=200, [20114-34978]) & HARD1 (C=100000, N=200, [20008-34991]) \\
    HARD2 (C=100000, N=200, [20009-34953]) & HARD3 (C=100000, N=200, [20033-34746]) \\
    HARD4 (C=100000, N=200, [20139-35000]) & HARD5 (C=100000, N=200, [20000-34955]) \\
    HARD6 (C=100000, N=200, [20050-34973]) & HARD7 (C=100000, N=200, [20017-34808]) \\
    HARD8 (C=100000, N=200, [20091-34992]) & HARD9 (C=100000, N=200, [20081-34991]) \\
    \midrule
    \multicolumn{2}{c}{\textbf{Schwerin Set 3 -- 30 instances}} \\
    \midrule
    Schwerin1\_BPP1 (C=1000, N=100, [150-200]) & Schwerin1\_BPP10 (C=1000, N=100, [150-200]) \\
    Schwerin1\_BPP11 (C=1000, N=100, [150-200]) & Schwerin1\_BPP12 (C=1000, N=100, [150-200]) \\
    Schwerin1\_BPP13 (C=1000, N=100, [150-199]) & Schwerin1\_BPP14 (C=1000, N=100, [150-200]) \\
    Schwerin1\_BPP15 (C=1000, N=100, [150-199]) & Schwerin1\_BPP2 (C=1000, N=100, [150-200]) \\
    Schwerin1\_BPP3 (C=1000, N=100, [150-200]) & Schwerin1\_BPP4 (C=1000, N=100, [150-200]) \\
    Schwerin1\_BPP5 (C=1000, N=100, [150-200]) & Schwerin1\_BPP6 (C=1000, N=100, [150-200]) \\
    Schwerin1\_BPP7 (C=1000, N=100, [150-200]) & Schwerin1\_BPP8 (C=1000, N=100, [150-200]) \\
    Schwerin1\_BPP9 (C=1000, N=100, [150-200]) & Schwerin2\_BPP1 (C=1000, N=120, [151-200]) \\
    Schwerin2\_BPP10 (C=1000, N=120, [150-199]) & Schwerin2\_BPP11 (C=1000, N=120, [150-200]) \\
    Schwerin2\_BPP12 (C=1000, N=120, [150-200]) & Schwerin2\_BPP13 (C=1000, N=120, [150-200]) \\
    Schwerin2\_BPP14 (C=1000, N=120, [150-200]) & Schwerin2\_BPP15 (C=1000, N=120, [151-200]) \\
    Schwerin2\_BPP2 (C=1000, N=120, [150-200]) & Schwerin2\_BPP3 (C=1000, N=120, [150-200]) \\
    Schwerin2\_BPP4 (C=1000, N=120, [150-200]) & Schwerin2\_BPP5 (C=1000, N=120, [150-200]) \\
    Schwerin2\_BPP6 (C=1000, N=120, [150-200]) & Schwerin2\_BPP7 (C=1000, N=120, [151-200]) \\
    Schwerin2\_BPP8 (C=1000, N=120, [150-200]) & Schwerin2\_BPP9 (C=1000, N=120, [150-200]) \\
    \midrule
    \multicolumn{2}{c}{\textbf{Falkenauer U (Uniform) -- 24 instances}} \\
    \midrule
    Falkenauer\_u1000\_00 (C=150, N=1000, [20-100]) & Falkenauer\_u1000\_01 (C=150, N=1000, [20-100]) \\
    Falkenauer\_u1000\_02 (C=150, N=1000, [20-100]) & Falkenauer\_u1000\_03 (C=150, N=1000, [20-100]) \\
    Falkenauer\_u1000\_04 (C=150, N=1000, [20-100]) & Falkenauer\_u1000\_05 (C=150, N=1000, [20-100]) \\
    Falkenauer\_u120\_00 (C=150, N=120, [20-98]) & Falkenauer\_u120\_01 (C=150, N=120, [20-100]) \\
    Falkenauer\_u120\_02 (C=150, N=120, [20-100]) & Falkenauer\_u120\_03 (C=150, N=120, [20-100]) \\
    Falkenauer\_u120\_04 (C=150, N=120, [20-99]) & Falkenauer\_u120\_05 (C=150, N=120, [20-100]) \\
    Falkenauer\_u250\_00 (C=150, N=250, [20-100]) & Falkenauer\_u250\_01 (C=150, N=250, [20-100]) \\
    Falkenauer\_u250\_02 (C=150, N=250, [20-100]) & Falkenauer\_u250\_03 (C=150, N=250, [20-100]) \\
    Falkenauer\_u250\_04 (C=150, N=250, [20-100]) & Falkenauer\_u250\_05 (C=150, N=250, [20-100]) \\
    Falkenauer\_u500\_00 (C=150, N=500, [20-100]) & Falkenauer\_u500\_01 (C=150, N=500, [20-100]) \\
    Falkenauer\_u500\_02 (C=150, N=500, [20-100]) & Falkenauer\_u500\_03 (C=150, N=500, [20-100]) \\
    Falkenauer\_u500\_04 (C=150, N=500, [21-100]) & Falkenauer\_u500\_05 (C=150, N=500, [20-100]) \\
    \midrule
    \multicolumn{2}{c}{\textbf{Falkenauer T (Triplets) -- 24 instances}} \\
    \midrule
    Falkenauer\_t120\_00 (C=1000, N=120, [250-497]) & Falkenauer\_t120\_01 (C=1000, N=120, [250-498]) \\
    Falkenauer\_t120\_02 (C=1000, N=120, [250-499]) & Falkenauer\_t120\_03 (C=1000, N=120, [250-499]) \\
    Falkenauer\_t120\_04 (C=1000, N=120, [250-499]) & Falkenauer\_t120\_05 (C=1000, N=120, [250-499]) \\
    Falkenauer\_t249\_00 (C=1000, N=249, [250-498]) & Falkenauer\_t249\_01 (C=1000, N=249, [250-499]) \\
    Falkenauer\_t249\_02 (C=1000, N=249, [250-496]) & Falkenauer\_t249\_03 (C=1000, N=249, [250-499]) \\
    Falkenauer\_t249\_04 (C=1000, N=249, [250-499]) & Falkenauer\_t249\_05 (C=1000, N=249, [250-499]) \\
    Falkenauer\_t501\_00 (C=1000, N=501, [250-498]) & Falkenauer\_t501\_01 (C=1000, N=501, [250-498]) \\
    Falkenauer\_t501\_02 (C=1000, N=501, [250-499]) & Falkenauer\_t501\_03 (C=1000, N=501, [250-499]) \\
    Falkenauer\_t501\_04 (C=1000, N=501, [250-499]) & Falkenauer\_t501\_05 (C=1000, N=501, [250-498]) \\
    Falkenauer\_t60\_00 (C=1000, N=60, [251-495]) & Falkenauer\_t60\_01 (C=1000, N=60, [251-475]) \\
    Falkenauer\_t60\_02 (C=1000, N=60, [250-498]) & Falkenauer\_t60\_03 (C=1000, N=60, [250-495]) \\
    Falkenauer\_t60\_04 (C=1000, N=60, [250-498]) & Falkenauer\_t60\_05 (C=1000, N=60, [250-496]) \\
    \bottomrule
  \end{tabular}}
\end{table}

\subsection{Flowshop}
We utilize standard Flowshop Scheduling benchmarks VRF~\cite{VRFBench:vallada2015new}. All instances aim to minimize the maximum completion time (Makespan).

\subsubsection{Test Sets}
We evaluate on four test sets categorized by job count (20, 40, 60, 100), encompassing various machine configurations. We report these sets as VRF20, VRF40, VRF60, and VRF100, while the underlying instance filenames use the \texttt{VFR} prefix. Table~\ref{tab:fsp-test-all} summarizes the test instances.

\begin{table}[t]
  \centering
  \caption{Combined Flowshop test instances (VRF20, VRF40, VRF60, and VRF100).}
  \label{tab:fsp-test-all}
  \scriptsize
  \resizebox{\linewidth}{!}{
  \begin{tabular}{@{}llll@{}}
    \toprule
    \multicolumn{2}{c}{\textbf{VRF20 (20 Jobs)}} & \multicolumn{2}{c}{\textbf{VRF40 (40 Jobs)}} \\
    \cmidrule(r){1-2} \cmidrule(l){3-4}
    \textbf{Instance Group} & \textbf{Count} & \textbf{Instance Group} & \textbf{Count} \\
    \midrule
    VFR20\_5\_\{1..10\} & 10 & VFR40\_5\_\{1..10\} & 10 \\
    VFR20\_10\_\{1..10\} & 10 & VFR40\_10\_\{1..10\} & 10 \\
    VFR20\_15\_\{1..10\} & 10 & VFR40\_15\_\{1..10\} & 10 \\
    VFR20\_20\_\{1..10\} & 10 & VFR40\_20\_\{1..10\} & 10 \\
    \midrule
    \multicolumn{2}{c}{\textbf{VRF60 (60 Jobs)}} & \multicolumn{2}{c}{\textbf{VRF100 (100 Jobs)}} \\
    \cmidrule(r){1-2} \cmidrule(l){3-4}
    \textbf{Instance Group} & \textbf{Count} & \textbf{Instance Group} & \textbf{Count} \\
    \midrule
    VFR60\_5\_\{1..10\} & 10 & VFR100\_20\_\{1..10\} & 10 \\
    VFR60\_10\_\{1..10\} & 10 & VFR100\_40\_\{1..10\} & 10 \\
    VFR60\_15\_\{1..10\} & 10 & VFR100\_60\_\{1..10\} & 10 \\
    VFR60\_20\_\{1..10\} & 10 & & \\
    \bottomrule
  \end{tabular}}
\end{table}

\subsection{CVRP}
The Capacitated Vehicle Routing Problem (CVRP) experiments utilize classical benchmark instances from the literature, including the sets by Augerat~\cite{CVRPDA:augerat1995computational} and Christofides~\cite{CVRPCMT:christofides1979vehicle}. These instances span a wide range of customer distributions, service node counts, and vehicle capacities.

\subsubsection{Test Sets}
Evaluation is performed on four standard benchmark sets. In the instance
naming convention (e.g., A-n32-k5), the number following ``n'' denotes the
total number of nodes, including one depot, while ``k'' denotes the number of
available vehicles. Thus, the number of service nodes is $n-1$. These sets
cover varied spatial distributions and demand/capacity ratios.

\begin{table}[t]
  \centering
  \caption{CVRP Test Sets overview.}
  \label{tab:cvrp-test}
  \scriptsize
  \begin{tabular}{@{}llcc@{}}
    \toprule
    \textbf{Set} & \textbf{Source} & \textbf{Service Nodes} & \textbf{Vehicle Capacity} \\
    \midrule
    Set A & Augerat et al. (1995) & 31--79 & 100 \\
    Set B & Augerat et al. (1995) & 30--77 & 100 \\
    Set P & Augerat et al. (1995) & 15--100 & 40--350 \\
    CMT & Christofides (1979) & 50--199 & 140--200 \\
    \bottomrule
  \end{tabular}
\end{table}

\begin{table}[t]
  \centering
  \caption{Complete list of the 88 CVRP test instances.}
  \label{tab:cvrp-test-all}
  \scriptsize
  \renewcommand{\arraystretch}{1.05}
  \begin{tabular}{@{}llll@{}}
    \toprule
    \textbf{A (27)} & \textbf{B (23)} & \textbf{P (24)} & \textbf{CMT (14)} \\
    \midrule
    A-n32-k5  & B-n31-k5  & P-n16-k8   & CMT1  \\
    A-n33-k5  & B-n34-k5  & P-n19-k2   & CMT2  \\
    A-n33-k6  & B-n35-k5  & P-n20-k2   & CMT3  \\
    A-n34-k5  & B-n38-k6  & P-n21-k2   & CMT4  \\
    A-n36-k5  & B-n39-k5  & P-n22-k2   & CMT5  \\
    A-n37-k5  & B-n41-k6  & P-n22-k8   & CMT6  \\
    A-n37-k6  & B-n43-k6  & P-n23-k8   & CMT7  \\
    A-n38-k5  & B-n44-k7  & P-n40-k5   & CMT8  \\
    A-n39-k5  & B-n45-k5  & P-n45-k5   & CMT9  \\
    A-n39-k6  & B-n45-k6  & P-n50-k7   & CMT10 \\
    A-n44-k6  & B-n50-k7  & P-n50-k8   & CMT11 \\
    A-n45-k6  & B-n50-k8  & P-n50-k10  & CMT12 \\
    A-n45-k7  & B-n51-k7  & P-n51-k10  & CMT13 \\
    A-n46-k7  & B-n52-k7  & P-n55-k7   & CMT14 \\
    A-n48-k7  & B-n56-k7  & P-n55-k8   &       \\
    A-n53-k7  & B-n57-k7  & P-n55-k10  &       \\
    A-n54-k7  & B-n57-k9  & P-n55-k15  &       \\
    A-n55-k9  & B-n63-k10 & P-n60-k10  &       \\
    A-n60-k9  & B-n64-k9  & P-n60-k15  &       \\
    A-n61-k9  & B-n66-k9  & P-n65-k10  &       \\
    A-n62-k8  & B-n67-k10 & P-n70-k10  &       \\
    A-n63-k9  & B-n68-k9  & P-n76-k4   &       \\
    A-n63-k10 & B-n78-k10 & P-n76-k5   &       \\
    A-n64-k9  &            & P-n101-k4  &       \\
    A-n65-k9  &            &            &       \\
    A-n69-k9  &            &            &       \\
    A-n80-k10 &            &            &       \\
    \bottomrule
  \end{tabular}
\end{table}

%% file: Sections/AppendixDetail/SHHDetail.tex

\section{Selection Hyper-Heuristic Framework}
\label{ap:SSH}

\subsection{Introduction of SHH}
Selection Hyper-heuristics (SHH) represent a class of high-level search methodologies designed to automate the selection and application of problem-specific heuristics for solving complex combinatorial optimization problems. As established in a comprehensive survey~\cite{HH2:drake2020recent}, the fundamental architecture of an SHH framework is defined by the "Domain Barrier." This conceptual separation decouples the high-level selection logic from the domain-specific data structures and Low-Level Heuristics (LLHs), ensuring that the high-level strategy remains problem-agnostic and transferable across different domains.

The SHH framework typically consists of two primary algorithmic components operating above the domain barrier:
\begin{itemize}
    \item \textbf{Selection Mechanism}: This component is responsible for choosing the most appropriate LLH from the available pool at each decision point. Modern selection mechanisms often utilize online learning, reinforcement learning, or statistical indicators to adaptively prioritize heuristics that demonstrate superior performance on the current instance or search phase.
    \item \textbf{Move Acceptance}: Once an LLH is applied and a new candidate solution is generated, the Move Acceptance component determines whether to accept the move. This ranges from simple deterministic rules (e.g., Only Improving) to probabilistic criteria inspired by meta-heuristics (e.g., the Metropolis acceptance criterion used in simulated annealing) to allow escape from local optima.
\end{itemize}

We select SHH as our research framework because its architectural design provides a natural environment for evaluating and optimizing multiple heuristics simultaneously. In this framework, a high-level strategy functions as a controller that dynamically selects and applies operators from a pool of pre-defined Low-Level Heuristics (LLHs) based on the current problem state and \textit{feedback information} (e.g., historical performance, objective value improvements) passed across the domain barrier. The existence of this LLH pool makes SHH naturally compatible with the objectives of Multi-Heuristic LLM-AHD, as it allows the LLM to focus on co-evolving a diverse ensemble of operators rather than a single isolated heuristic. Furthermore, the dynamic scheduling managed by the high-level strategy leads to complex, synergistic relationships among the LLHs, where the utility of a specific operator is often conditional on the previous search history.

Algorithm~\ref{alg:generic-shh} summarizes the generic execution procedure. At each decision point, the selection mechanism chooses an LLH using the current search state and historical feedback. The selected LLH produces a candidate solution, which is then evaluated by the move acceptance mechanism. The resulting performance information is retained to guide subsequent selections.

\begin{algorithm}[t]
\caption{Generic Selection Hyper-Heuristic Framework}
\label{alg:generic-shh}
\begin{algorithmic}[1]
\STATE \textbf{Input:} Instance $I$ and LLH pool $\mathcal{H}$
\STATE \textbf{Input:} Selection mechanism $\mathcal{S}$, move acceptance $\mathcal{A}$, and termination rule $\mathcal{T}$
\STATE $x \leftarrow \textsc{Initialize}(I)$; $x^* \leftarrow x$; history $\mathcal{B} \leftarrow \emptyset$
\WHILE{$\mathcal{T}$ is not satisfied}
    \STATE $z \leftarrow \textsc{ObserveState}(x,x^*,\mathcal{B})$
    \STATE $h \leftarrow \mathcal{S}(\mathcal{H},z,\mathcal{B})$
    \STATE $x' \leftarrow \textsc{ApplyLLH}(h,x,I)$
    \STATE $b \leftarrow \mathcal{A}(x,x',z,\mathcal{B})$
    \IF{$b$}
        \STATE $x \leftarrow x'$
    \ENDIF
    \IF{$f(x)<f(x^*)$}
        \STATE $x^* \leftarrow x$
    \ENDIF
    \STATE $\mathcal{B} \leftarrow \textsc{UpdateFeedback}(\mathcal{B},h,x',b,x,x^*)$
\ENDWHILE
\STATE \textbf{return} $x^*$
\end{algorithmic}
\end{algorithm}

\subsection{Design Principle}

We implement our framework based on the standard HyFlex interface established in the CHeSC 2011~\cite{Hyflex:ochoa2012hyflex}. While HyFlex provides a structural foundation, existing SHH frameworks typically exhibit complex architectural structures, presenting significant challenges to the direct application of LLM-AHD. For example, these heuristics frequently manipulate domain-specific solution representations directly, lacking safe data manipulation interfaces. Crucially, while such structural inconsistencies are effectively masked when combined with traditional high-level search strategies, they become critical issues when employing LLM-AHD to directly optimize LLHs. Consequently, LLM-generated code is highly prone to hallucinations, resulting in invocations of non-existent methods or invalid manipulations of structured data. To systematically address these challenges and enable reliable LLM-driven heuristic optimization within SHH frameworks, we propose three main design principles that transform a traditional hyper-heuristic framework into an LLM-compatible framework.

\textbf{Principle 1: Unified LLH Invocation Interface.}
To facilitate the simultaneous evolution of multiple heterogeneous heuristics, we establish a standardized interaction protocol. In standard implementations, distinct operator types (e.g., mutation versus local search) often require divergent function signatures and variable naming conventions. This inconsistency frequently induces hallucinations, where the LLM erroneously generates code with incorrect function signatures or mismatched parameter definitions. We address this by enforcing a homogenized invocation schema for all Low-Level Heuristics (LLHs). This unification ensures that the LLM interacts with a consistent API surface regardless of the specific operator type being evolved, thereby effectively preventing syntax errors derived from signature mismatches during the iterative co-evolutionary process.

\textbf{Principle 2: Error Prevention via Validated Operation Encapsulation.}
Complex combinatorial solvers often rely on complex problem-specific data structures that are error-prone when directly manipulated by LLM-generated code. Naive exposure of these internal representations can lead to a cascade of runtime failures—invalid state transitions, constraint violations, or inconsistent objective updates. We address this by encapsulating all low-level data manipulations into validated operation interfaces that enforce domain-specific boundary checks. By restricting the LLM's interaction to this validated API surface, we significantly mitigate common error patterns (e.g., out-of-bounds access, solution corruption).

\textbf{Principle 3: Domain-Specific Algorithmic Library.}
In traditional automated heuristic design approaches, LLMs are often tasked with implementing fundamental operations from scratch, which distracts their attention from high-level strategic reasoning and consumes valuable context budget. To address this inefficiency, we curate and modularize domain-specific fundamental heuristics (e.g., greedy insert, nearest neighbor for TSP) into a structured algorithmic library. By providing these pre-validated building blocks, the framework liberates the LLM from the burden of reimplementing basic operators for every generated heuristic. This design shift enables the model to focus its generative capacity on novel heuristic designs.

\subsection{Implementation Interfaces}

The controller accesses every unary LLH through
\texttt{apply\_heuristic(heuristic\_id, from\_index, to\_index)} and every
crossover through
\texttt{apply\_heuristic\_with\_parents(heuristic\_id, parent1\_index,
parent2\_index, to\_index)}. These calls copy or combine solutions in managed
memory, dispatch the selected original or evolved implementation, recompute
the objective when required, validate feasibility, and return the objective
value stored at \texttt{to\_index}.

The code generated for unary LLHs has the domain-level contract
\texttt{def method(self, solution: SolutionType) -> None}. It modifies the
provided solution in place and cannot directly manage controller memory.
Crossover code receives two parent solutions and produces the destination
solution through the validated domain adapter. Consequently, the LLM-facing
implementation contract remains domain-specific enough to expose useful
features, while the controller-facing invocation contract is identical across
TSP, BPP, Flowshop, and CVRP.

\subsection{Low-Level Heuristics}
\label{app:lhh}

We follow the HyFlex LLH taxonomy: \textbf{Mutation}, \textbf{Ruin-Recreate}, \textbf{Local Search}, and \textbf{Crossover}. A hyper-heuristic is a high-level controller that selects or sequences LLHs across these types, so it is not itself an LLH type. The details are shown in Table~\ref{tab:heuristic_config}.

\textbf{Mutation.}
Mutation heuristics perturb or reorder parts of a single solution to generate a new search state. Their changes can range from small edits, such as swapping two elements, to stronger randomization of a subsequence or complete representation. They primarily provide diversification while preserving the domain's feasibility requirements.

\textbf{Ruin-Recreate.}
Ruin-Recreate heuristics remove a selected portion of the current solution and reconstruct the missing structure using a domain-specific insertion or packing rule. By combining a disruptive removal stage with a guided rebuilding stage, they can move the search beyond neighborhoods reachable through simple mutations.

\textbf{Local Search.}
Local Search heuristics repeatedly examine a structured neighborhood and apply improving moves to refine the current solution. Different operators vary in neighborhood definition, move evaluation, and stopping policy, providing complementary forms of intensification at different computational costs.

\textbf{Crossover.}
Crossover heuristics construct a new solution from two parent solutions by preserving and recombining useful structural information from both. Domain-specific repair or representation-aware recombination ensures that the resulting offspring remains valid. These operators allow the framework to exploit information distributed across multiple search trajectories.

\definecolor{lightgray}{gray}{0.95}

\begin{table*}[t]
\centering
\caption{Detailed configuration of heuristic operators for TSP and BPP.}
\label{tab:heuristic_config}
\scriptsize
\renewcommand{\arraystretch}{1.15}
\begin{tabular}{@{}>{\raggedright\arraybackslash}p{0.25\textwidth}
                    >{\raggedright\arraybackslash}p{0.15\textwidth}
                    >{\raggedright\arraybackslash}p{0.50\textwidth}@{}}
\toprule
\textbf{Operator Name (ID)} & \textbf{Type} & \textbf{Description} \\
\midrule
\rowcolor{lightgray} \multicolumn{3}{l}{\textbf{Traveling Salesman Problem (TSP)}} \\
\midrule
random\_reinsertion (0) & Mutation & Removes one city and reinserts it at a random position to preserve feasibility. \\
swap\_two (1) & Mutation & Swaps two randomly selected cities. \\
shuffle (2) & Mutation & Randomly shuffles the entire permutation. \\
shuffle\_subsequence (3) & Mutation & Shuffles a randomly chosen subset of positions. \\
n\_opt\_move (4) & Mutation & Applies a sequence of 2-opt flips to rewire edges. \\
iterated\_greedy (5) & Ruin-Recreate & Removes a fraction of cities and reinserts them greedily. \\
two\_opt\_local\_search (6) & Local Search & 2-opt edge swaps with first-improvement policy. \\
best\_imp\_two\_opt (7) & Local Search & 2-opt edge swaps with best-improvement policy. \\
three\_opt\_local\_search (8) & Local Search & Uses 3-opt reconnections for complex route changes. \\
order\_crossover (9) & Crossover & Preserves a subsequence from one parent and fills remaining cities in order. \\
partially\_mapped (10) & Crossover & Exchanges segments and maps duplicates (PMX). \\
precedence\_preservative (11)& Crossover & Merges parent precedence constraints (PPC). \\
one\_point\_crossover (12) & Crossover & Splices two parents at one point and repairs duplicates. \\

\midrule
\rowcolor{lightgray} \multicolumn{3}{l}{\textbf{Bin Packing Problem (BPP)}} \\
\midrule
random\_piece\_move (0) & Mutation & Moves random pieces between bins. \\
swap\_pieces (3) & Mutation & Swaps pieces between two bins when feasible. \\
random\_bin\_merge (5) & Mutation & Merges contents of two bins if capacity allows. \\
remove\_repack\_rnd (1) & Ruin-Recreate & Removes random items and repacks using first-fit. \\
empty\_bins\_repack (2) & Ruin-Recreate & Empties selected bins and repacks their items. \\
best\_fit\_improvement (4) & Local Search & Moves items to better-fitting bins to reduce waste. \\
bin\_consolidation (6) & Local Search & Consolidates bins by merging compatible ones. \\
bin\_crossover (7) & Crossover & Combines bins from parents and packs remainder with first-fit. \\
\bottomrule
\end{tabular}
\end{table*}

\begin{table*}[t]
\centering
\caption{Detailed configuration of heuristic operators for Flowshop and CVRP (continued).}
\scriptsize
\renewcommand{\arraystretch}{1.15}
\begin{tabular}{@{}>{\raggedright\arraybackslash}p{0.25\textwidth}
                    >{\raggedright\arraybackslash}p{0.15\textwidth}
                    >{\raggedright\arraybackslash}p{0.50\textwidth}@{}}
\toprule
\textbf{Operator Name (ID)} & \textbf{Type} & \textbf{Description} \\
\midrule
\rowcolor{lightgray} \multicolumn{3}{l}{\textbf{Flowshop Scheduling}} \\
\midrule
random\_reinsertion (0) & Mutation & Removes one job and reinserts it at a random position. \\
swap\_two (1) & Mutation & Swaps two randomly selected jobs. \\
shuffle (2) & Mutation & Randomly shuffles the entire job permutation. \\
shuffle\_subsequence (3) & Mutation & Shuffles a randomly chosen subset of job positions. \\
use\_neh (4) & Mutation & Applies NEH heuristic using current permutation as initial. \\
iterated\_greedy (5) & Ruin-Recreate & Removes jobs and reinserts using best-position insertion. \\
deep\_iterated\_greedy (6) & Ruin-Recreate & Applies multiple rounds of iterated greedy. \\
local\_search (7) & Local Search & Job reinsertion with best-improvement policy. \\
first\_imp\_local\_search (8) & Local Search & Job reinsertion with first-improvement policy. \\
random\_local\_search (9) & Local Search & Random job reinsertion based on search depth. \\
rnd\_first\_imp\_ls (10) & Local Search & Random job selection with first-improvement reinsertion. \\
order\_crossover (11) & Crossover & Preserves subsequence and fills remainder in order. \\
precedence\_preservative (12)& Crossover & Merges parent precedence constraints. \\
partially\_mapped (13) & Crossover & Exchanges segments and uses mapping to resolve duplicates. \\
one\_point\_crossover (14) & Crossover & Splices parents at one point and repairs sequence. \\

\midrule
\rowcolor{lightgray} \multicolumn{3}{l}{\textbf{Capacitated VRP (CVRP)}} \\
\midrule
two\_opt (0) & Mutation & Reverses a segment within a single route. \\
or\_opt (1) & Mutation & Relocates a sequence of 2 customers within the same route. \\
shift\_mutate (5) & Mutation & Moves a customer to another route with random insertion. \\
location\_ruin\_recreate (2) & Ruin-Recreate & Spatial ruin around a random location with best-fit repair. \\
shift (3) & Local Search & Relocates customers between routes based on distance penalty. \\
two\_opt\_star (6) & Local Search & Exchanges route tails between two different routes. \\
geni (7) & Local Search & Generalized insertion heuristic (GENI) for non-adjacent insertion. \\
combine (4) & Crossover & Merges high-quality routes from parents. \\
\bottomrule
\end{tabular}
\end{table*}

%% file: Sections/AppendixDetail/ACODetail.tex

\section{Componentized Ant Colony Optimization Framework}
\label{ap:aco}

\subsection{Introduction of ACO}

Ant Colony Optimization (ACO) is a population-based metaheuristic inspired by the collective foraging behavior of ant colonies~\cite{ACO:dorigo1996ant}. A set of artificial ants incrementally constructs candidate solutions using pheromone information and problem-dependent heuristic information. High-quality solutions reinforce useful decisions through pheromone updates, thereby biasing later ants toward promising regions while preserving stochastic exploration.

A typical ACO iteration initializes the construction state of the ants, controls the search parameters, and repeatedly selects feasible actions until complete solutions are obtained. The constructed solutions may then be improved by local search. Finally, the framework assesses search progress, maintains high-quality solutions, and updates the global pheromone state for the next iteration. These stages provide the execution structure for the nine components described below.

\subsection{Design Principle}

\textbf{Principle 1: Fixed Component Interfaces.}
Each ACO component has a fixed function signature, input state, output format, and default implementation. Although the components operate at different stages and serve different purposes, the framework invokes each one at a predefined point in the search process. Consequently, an individual component can focus on its own algorithmic decision without reimplementing the complete ACO solver.

\textbf{Principle 2: Trusted Domain Kernel.}
Instance parsing, objective evaluation, feasible-action masking, constraint validation, and termination checks remain within a trusted domain kernel. The components control the ACO search decisions but cannot bypass the fundamental feasibility requirements of TSP or CVRP. This separation allows component implementations to concentrate on search strategy while the framework preserves valid solver execution.

\subsection{ACO Components}

The nine components are organized by their positions in the ACO workflow. H1--H3 initialize search information and control parameters; H4 and H9 govern solution construction; H5 improves completed solutions; and H6--H8 monitor search progress, maintain elite solutions, and perform global learning. Table~\ref{tab:aco-components} summarizes their individual responsibilities.

\begin{table*}[t]
\centering
\caption{Components in the componentized ACO framework.}
\label{tab:aco-components}
\scriptsize
\renewcommand{\arraystretch}{1.12}
\begin{tabular}{@{}>{\centering\arraybackslash}p{0.06\textwidth}
                    >{\raggedright\arraybackslash}p{0.28\textwidth}
                    >{\raggedright\arraybackslash}p{0.58\textwidth}@{}}
\toprule
\textbf{ID} & \textbf{Component} & \textbf{Responsibility} \\
\midrule
H1 & \texttt{build\_static\_heuristic} & Constructs static heuristic information from problem features before the search begins. \\
H2 & \texttt{control\_parameters} & Sets or adapts the principal ACO parameters according to the current search state. \\
H3 & \texttt{initialize\_ant\_starts} & Determines the starting positions and initial construction states of the ants. \\
H4 & \texttt{local\_pheromone\_update} & Applies local pheromone updates during incremental solution construction. \\
H5 & \texttt{improve\_solutions} & Applies local improvement to completed candidate solutions. \\
H6 & \texttt{detect\_stagnation} & Detects search stagnation from the optimization history and current pheromone state. \\
H7 & \texttt{manage\_solution\_archive} & Maintains high-quality solutions and associated archive information. \\
H8 & \texttt{global\_pheromone\_control} & Performs global pheromone updates and other iteration-level control operations. \\
H9 & \texttt{select\_actions} & Selects the next action from the feasible candidates using the transition information. \\
\bottomrule
\end{tabular}
\end{table*}

\textbf{Initialization and parameter control (H1--H3).}
H1 produces the static heuristic information once before search, H2 controls the principal parameters from the current search state, and H3 initializes the ants at the beginning of each construction phase. Together, these components determine the search conditions under which an iteration begins.

\textbf{Solution construction (H4 and H9).}
H9 selects the next action from the feasible candidates scored by the trusted kernel. After the selected actions are applied, H4 modifies the local pheromone state. Both components are called repeatedly while the ants incrementally construct complete solutions, providing decision making and immediate feedback within an iteration.

\textbf{Solution improvement (H5).}
H5 receives the completed candidate solutions and attempts to improve their objective values. The improved solutions are subsequently used in best-solution tracking and pheromone control, while problem-specific feasibility remains enforced by the domain kernel.

\textbf{Search monitoring and global learning (H6--H8).}
H6 determines whether the search is stagnating, H7 maintains an archive of high-quality solutions, and H8 updates the global pheromone state using the current solutions, best-so-far solution, archive, and stagnation information. These components carry search information across iterations and regulate longer-term search behavior.

\subsection{Implementation Provenance and Hook Interfaces}

The trusted numerical kernel was adapted from the Ant System solver used in
the public DeepACO implementation~\cite{ACO:ye2023deepaco}. We use neither
DeepACO's neural heuristic model nor its training procedure. The
componentization itself is our implementation: an initial six-hook
refactoring was extended with explicit stagnation detection, elite-archive
management, and action selection to obtain the nine-hook interface used in
the experiments.

The default hooks combine established ACO ideas with implementation-specific
control rules. H1 and H9 retain the classical visibility and stochastic
transition structure of Ant System~\cite{ACO:dorigo1996ant}; H4 uses an
ACS-style local update~\cite{ACO:dorigo1997acs}; H7 follows the
population/archive perspective of P-ACO~\cite{ACO:guntsch2002population};
and H8 combines rank-based or best-so-far reinforcement and bounded
restart-oriented control inspired by Rank-Based Ant System and
MAX--MIN Ant System~\cite{ACO:bullnheimer1999rank,ACO:stutzle2000maxmin}.
The adaptive schedules, multi-signal stagnation detector, archive bias,
candidate-feature corrections, and the particular combinations of these
mechanisms are custom policies rather than claims of a standard nine-component
ACO architecture.

The exact callable interfaces are shown below. \texttt{state} and other
dictionary arguments may contain optional information, but every generated
hook must preserve the listed positional contract and return structure.

\begin{tcolorbox}[colback=gray!3, colframe=gray!50,
title=Nine ACO Hook Interfaces, fonttitle=\bfseries]
\scriptsize
\begin{verbatim}
H1 build_static_heuristic(
    context, candidate_info, state=None) -> ndarray

H2 control_parameters(
    iteration, best_history, pheromone_stats,
    stagnation_info, context, state=None) -> dict

H3 initialize_ant_starts(
    n_ants, context, candidate_info, iteration,
    state=None) -> ndarray

H4 local_pheromone_update(
    pheromone, edge, construction_state, params,
    context, global_state=None) -> ndarray

H5 improve_solutions(
    solutions, costs, context, candidate_info, params,
    iteration, archive, state=None) -> (solutions, costs)

H6 detect_stagnation(
    iteration, best_history, pheromone, solutions, costs,
    context, state=None) -> dict

H7 manage_solution_archive(
    archive, new_solutions, new_costs, best_solution,
    best_cost, stagnation_info, params, iteration,
    context, state=None) -> list[(solution, cost)]

H8 global_pheromone_control(
    pheromone, solutions, costs, best_solution, best_cost,
    params, iteration, stagnation_info, archive, context,
    state=None) -> (pheromone, state)

H9 select_actions(
    scores, fallback_mask, construction_state, params,
    context, global_state=None) -> ndarray
\end{verbatim}
\end{tcolorbox}

Algorithm~\ref{alg:componentized-aco} shows where the nine components are invoked. Operations marked as kernel operations remain fixed, whereas H1--H9 correspond to the replaceable components listed in Table~\ref{tab:aco-components}.

\begin{algorithm}[t]
\caption{Componentized Ant Colony Optimization}
\label{alg:componentized-aco}
\begin{algorithmic}[1]
\STATE \textbf{Input:} Instance $I$, ants $M$, and iteration limit $T$
\STATE $\eta\leftarrow H_1(I)$; initialize $\tau$, best solution $x^*$, archive $\mathcal{A}$, and history $\mathcal{B}$
\FOR{$t=1,\ldots,T$}
    \STATE $\theta_t\leftarrow H_2(t,\mathcal{B})$; $X\leftarrow H_3(I,M,\theta_t)$
    \WHILE{some solution in $X$ is incomplete}
        \STATE $(Q,F)\leftarrow\textsc{KernelScores}(I,X,\tau,\eta,\theta_t)$
        \STATE $a\leftarrow H_9(Q,F,X)$; $X\leftarrow\textsc{KernelApply}(X,a)$
        \STATE $\tau\leftarrow H_4(\tau,X,a,\theta_t)$
    \ENDWHILE
    \STATE $X\leftarrow H_5(I,X,\theta_t)$; $c\leftarrow\textsc{KernelEvaluate}(I,X)$
    \STATE $x^*\leftarrow\textsc{UpdateBest}(x^*,X,c)$
    \STATE $s\leftarrow H_6(\mathcal{B},\tau,c)$; $\mathcal{A}\leftarrow H_7(\mathcal{A},X,c)$
    \STATE $\tau\leftarrow H_8(\tau,X,c,x^*,\mathcal{A},s,\theta_t)$
    \STATE $\mathcal{B}\leftarrow\textsc{UpdateHistory}(\mathcal{B},c,x^*,s)$
\ENDFOR
\STATE \textbf{return} $x^*$
\end{algorithmic}
\end{algorithm}

%% file: Sections/AppendixDetail/ExperimentDetail.tex
\section{Implementation and Experimental Details}
\label{app:implementation_experiments}

\subsection{Experimental Protocol and Search Budget}
\label{app:experimental_protocol}

Table~\ref{tab:common_experimental_config} reports the configuration shared
by the experiments in the main paper. A function evaluation (FE) denotes the
complete evaluation of one proposed component or jointly proposed pair on the
full training set. For MuEvo, this evaluation covers the Best, Initial, and
Secondary collaboration contexts under one shared per-candidate solver-run
budget. Each regular evolution round or pair-evolution round therefore
consumes exactly one FE; the contexts are not charged as separate FEs.
Reflection, code generation, static validation, and failed output parsing do
not consume additional FEs.

\begin{table}[t]
\centering
\caption{Common experimental configuration.}
\label{tab:common_experimental_config}
\scriptsize
\renewcommand{\arraystretch}{1.12}
\begin{tabular}{@{}p{0.57\columnwidth}p{0.34\columnwidth}@{}}
\toprule
\textbf{Parameter} & \textbf{Value} \\
\midrule
Primary LLM & DeepSeek-V4-Flash \\
Generator temperature & 0.8 \\
Generator output limit & 6,144 tokens \\
Independent AHD runs per method & 3 \\
MuEvo, Single, and Batch main-search budget & 100 FEs \\
Split main-search budget & 100 FEs per target; 300 FEs total \\
Component-probing budget & 25 FEs per valid component \\
Probe budget accounting & Excluded from the main-search budget \\
Within-component population size & 4 \\
Mutation sampling fraction & 0.3 \\
Training solver runs per instance and FE & 10 \\
Test solver runs per instance & 10 \\
Test time limit per solver run & 300 seconds \\
Training controller for SHH & ADAPHH \\
Hardware & Two AMD EPYC 9754 CPUs; 1,024 GB RAM \\
\bottomrule
\end{tabular}
\end{table}

All methods use the same 25-FE-per-component preliminary budget. MuEvo and the
Single and Batch adaptations then use 100 main-search FEs. The Split
adaptation instead runs three independent 100-FE searches over the same Top-3
components used by Batch, for 300 main-search FEs in total. This allocation is
intentionally favorable to Split rather than a claim of equal total
computation; the exact target selection and accounting are detailed in
Appendix~E. All final tables average the three independent AHD runs; each
resulting solver is evaluated with independent solver seeds.

\subsection{MuEvo Hyperparameters}
\label{app:muevo_hyperparameters}

Table~\ref{tab:muevo_hyperparameters} lists the parameters corresponding to
the mechanisms in Section~3 of the main paper. These values are fixed across
the reported domains and are not tuned separately on the test sets.

\begin{table*}[t]
\centering
\caption{MuEvo hyperparameters used in the reported experiments.}
\label{tab:muevo_hyperparameters}
\scriptsize
\renewcommand{\arraystretch}{1.12}
\begin{tabular}{@{}p{0.22\textwidth}p{0.43\textwidth}p{0.25\textwidth}@{}}
\toprule
\textbf{Mechanism} & \textbf{Parameter} & \textbf{Value} \\
\midrule
Component probing
& Active components after probing, $K_A$ & 3 \\
\midrule
Dynamic component lifecycle
& Global-stagnation threshold for activation, $\tau_{\mathrm{stag}}$ & 1 round \\
& Minimum interval between activations, $\tau_{\mathrm{int}}$ & 2 rounds \\
& Consecutive gate rejections before Active-to-Inactive transition, $\tau_{\mathrm{reject}}$ & 2 \\
& Inactive reactivation cooldown, $L_{\mathrm{cool}}$ & 3 rounds \\
& Active-to-Inactive stagnation threshold & 4 selections \\
& Force-select a newly activated component & Yes, for one round \\
\midrule
Multi-Ensemble Evaluation
& Context weights (Best, Initial, Secondary) & $(0.6,0.2,0.2)$ \\
& Solver-run allocation (Best, Initial, Secondary) & $(4,3,3)$ of 10 \\
& Worst-context coefficient, $\lambda_{\min}$ & 0.2 \\
& Cross-context variation coefficient, $\lambda_{\mathrm{var}}$ & 0.1 \\
& Acceptance threshold for $R_i$ & 0 \\
& Require improvement in the Best context & Yes \\
& Context evaluation order & Best, Initial, Secondary \\
& Ensemble archive size & 8 \\
& Minimum archive diversity for Secondary context & 0.25 \\
\midrule
Cross-Component Information Sharing
& Maximum stored replacement events & 200 \\
& Recent events included per target & 6 \\
& Minimum events before updating an LLM relation summary & 2 \\
& Relation-summary update interval / evidence window & 2 rounds / 8 events \\
& Maximum relation-context / summary lengths & 2,000 / 1,600 characters \\
& Relation-summary temperature / output limit & 0.8 / 900 tokens \\
\midrule
Relation-Guided Pair Evolution
& Pair-evolution interval, $q$ & 2 regular rounds \\
& First eligible pair round / maximum pair events & 2 / 5 \\
& Conservative remaining-budget guard & 3 FEs \\
& Pair failure penalty / accepted-pair bonus & 0.30 / 0.20 \\
& Repeated failures before pair cooldown / cooldown length & 2 / 3 rounds \\
& Pair-generation temperature / output limit & 0.8 / 4,096 tokens \\
& Maximum format-repair requests & 1 \\
& Maximum preflight-repair requests & 2 \\
\midrule
Adaptive Budget Allocation
& Improvement-rate weight, $w_I$ & 0.40 \\
& Fitness-variance weight, $w_V$ & 0.30 \\
& Stagnation weight / saturation horizon, $w_S,H_z$ & 0.05 / 5 selections \\
& Recent-reward weight / history window, $w_R,H_r$ & 0.25 / 5 selections \\
& Effective allocation-frequency penalty, $w_\rho$ & 0.40 \\
& Non-positive-reward streak penalty / saturation & 0.20 / 3 selections \\
\bottomrule
\end{tabular}
\end{table*}

\subsubsection{Dynamic Component Management}

The lifecycle contains exactly two component states: Active and Inactive. The
top three probed components are initialized as Active, and every remaining
valid component is initialized as Inactive. An Inactive component becomes
eligible for activation when global stagnation reaches one round and at least
two rounds have elapsed since the previous activation; the selected component
is promoted to Active and force-selected for one round. An Active component
is moved to Inactive after two consecutive rejected Multi-Ensemble Evaluations
and cannot be reactivated during the following three-round cooldown. It can
also be moved to Inactive after four selections without improvement, provided
another Active component remains.

Inactive activation candidates are ordered by fewer consecutive gate
rejections, higher uncertainty and probing scores, larger exponentially
smoothed Best-context gains, fewer previous activations, and component ID.
The exponential moving average uses coefficient 0.35. Components in cooldown
are excluded from activation. The reported lifecycle does not introduce any
additional component state.

\subsubsection{Multi-Ensemble Evaluation}

Best, Initial, and Secondary contexts receive weights 0.6, 0.2, and 0.2.
The ten solver runs assigned to one candidate on each training instance are
shared across the three contexts: four runs use the Best context, three use
the Initial context, and three use the Secondary context. These ten runs
together constitute one FE, rather than ten run-level FEs or three
context-level FEs. The three resulting deltas are aggregated using Eq.~(5) of
the main paper with $\lambda_{\min}=0.2$ and
$\lambda_{\mathrm{var}}=0.1$. Acceptance requires both
$\Delta_i^{\mathrm{Best}}>0$ and $R_i>0$. The Secondary context is selected
from an archive of at most eight ensembles, subject to a minimum diversity of
0.25 from the Best context.

\subsubsection{Relation Memory and Pair Evolution}

Relation memory retains at most 200 accepted or rejected replacement events.
For a selected target, the six most recent relevant events are combined with
its interaction-graph neighborhood. The LLM-based interaction summary is
updated every two rounds after at least two events are available and uses at
most eight events. The SHH interaction graph starts without hand-coded
relation priors and is learned from single- and pair-replacement outcomes;
the componentized ACO graph may additionally be initialized with structural
dependencies between hooks.

Pair evolution is attempted every two regular rounds, beginning with the
second round, while at least three FEs remain. Eligible pairs are ranked by
their structural or empirically learned interaction score. For pair
$(i,j)$, the implementation adds 0.20 for each previously accepted joint
replacement, subtracts 0.30 for each failed attempt and an additional 0.15
per consecutive failure, and excludes a pair for three rounds after two
consecutive failures. Ties are resolved by the graph score and then the
lexicographic component IDs. A pair is accepted only as a joint replacement
through the same Multi-Ensemble Evaluation rule used for single components,
and the complete pair evaluation consumes one FE. The three-FE condition is a
conservative scheduling guard that preserves budget for subsequent evolution;
it is not the evaluation cost of the pair.

\subsubsection{Adaptive Budget Allocation}

The implementation uses the quantities in Eq.~(6) of the main paper with
$w_I=0.40$, $w_V=0.30$, $w_S=0.05$, $H_z=5$, $w_R=0.25$, and $H_r=5$.
Under-exploration contributes $0.30(1-\rho_i)$ and the explicit
over-selection term contributes $-0.10\rho_i$; because the additive constant
does not affect ranking, these are equivalent to an effective allocation
penalty $w_\rho=0.40$. A supplementary penalty of at most 0.20 is applied
when the most recent three rewards are non-positive. Equal scores are broken
by the smaller component ID. Lifecycle-forced activation takes precedence
over this ranking for one round.

\subsection{Cross-Controller Generalization Details}
\label{app:cross_controller_details}

Table~\ref{tab:generalization} reports the complete TSP and BPP results used
in the cross-controller analysis. ``Ori'' denotes the original human-designed
LLH ensemble, and ``Evo'' denotes the same controller using the LLH ensemble
evolved with ADAPHH; no controller-specific retraining is performed.

\begin{table*}[t]
\centering
\caption{Cross-controller generalization. Values report average cost or bins
and optimality gap (\%) for the original and MuEvo-evolved LLH ensembles.}
\label{tab:generalization}
\resizebox{\textwidth}{!}{%
\begin{tabular}{l|cccc|cccc}
\hline
\textbf{Domain} & \multicolumn{4}{c|}{\textbf{TSP}} & \multicolumn{4}{c}{\textbf{Bin Packing}} \\
\hline
\textbf{Dataset} & \textbf{TSP-SS} & \textbf{TSP-S} & \textbf{TSP-M} & \textbf{TSP-L} & \textbf{Falkenauer\_U} & \textbf{Falkenauer\_T} & \textbf{Scholl\_3} & \textbf{Schwerin} \\
\hline
\textbf{Method} & Cost (Gap) $\downarrow$ & Cost (Gap) $\downarrow$ & Cost (Gap) $\downarrow$ & Cost (Gap) $\downarrow$ & Bins (Gap) $\downarrow$ & Bins (Gap) $\downarrow$ & Bins (Gap) $\downarrow$ & Bins (Gap) $\downarrow$ \\
\hline
ADAPHH-Ori & 30211.7 (1.875\%) & 26504.9 (4.756\%) & 92521.4 (5.765\%) & 44680.6 (7.157\%) & 190.667 (0.754\%) & 82.9958 (6.484\%) & 58.1800 (3.530\%) & 20.3467 (2.166\%) \\
ADAPHH-Evo & \textbf{29935.8 (0.651\%)} & \textbf{25815.0 (2.141\%)} & \textbf{90672.5 (3.657\%)} & \textbf{43909.0 (5.237\%)} & \textbf{189.017 (0.051\%)} & \textbf{78.4667 (2.159\%)} & \textbf{56.4000 (0.360\%)} & \textbf{20.0067 (0.394\%)} \\
\hline
HAHA-Ori & 30081.4 (1.348\%) & 26327.7 (4.201\%) & 92501.6 (5.795\%) & 45148.7 (8.170\%) & 190.104 (0.491\%) & 80.8667 (4.117\%) & 57.2700 (1.910\%) & 20.1400 (1.095\%) \\
HAHA-Evo & \textbf{29948.0 (0.648\%)} & \textbf{25902.3 (2.402\%)} & \textbf{91948.4 (5.251\%)} & \textbf{44878.9 (8.067\%)} & \textbf{189.496 (0.227\%)} & \textbf{78.5292 (2.343\%)} & \textbf{56.4200 (0.668\%)} & \textbf{19.9667 (0.159\%)} \\
\hline
NAHH-Ori & 30038.2 (1.137\%) & 26751.0 (5.679\%) & 95271.6 (8.944\%) & 46179.8 (11.096\%) & 190.054 (0.362\%) & 80.3167 (3.580\%) & 56.6300 (0.771\%) & 20.0300 (0.532\%) \\
NAHH-Evo & \textbf{29930.4 (0.581\%)} & \textbf{26008.0 (2.747\%)} & \textbf{92108.6 (5.337\%)} & \textbf{45925.2 (10.307\%)} & \textbf{189.371 (0.160\%)} & \textbf{78.6625 (2.423\%)} & \textbf{56.4000 (0.361\%)} & \textbf{20.0067 (0.402\%)} \\
\hline
VNSTW-Ori & 30137.3 (1.470\%) & 26567.5 (4.905\%) & 93985.0 (9.134\%) & 49204.5 (18.149\%) & 191.238 (0.674\%) & 80.1958 (3.488\%) & 56.6700 (0.844\%) & 20.1467 (1.126\%) \\
VNSTW-Evo & \textbf{29956.8 (0.701\%)} & \textbf{26076.5 (3.253\%)} & \textbf{92168.9 (5.591\%)} & \textbf{44443.1 (6.741\%)} & \textbf{189.317 (0.154\%)} & \textbf{78.5042 (2.328\%)} & \textbf{56.4000 (0.361\%)} & \textbf{20.1000 (0.897\%)} \\
\hline
\end{tabular}%
}
\end{table*}

\subsection{Sample Evolved Heuristic Code}

Below is an excerpt from an evolved heuristic (H5: Iterated Greedy) generated by MuEvo for TSP:

\begin{tcolorbox}[colback=gray!3, colframe=gray!50, title=Evolved Heuristic: Iterated Greedy (TSP), fonttitle=\bfseries]
\scriptsize
\begin{verbatim}
def _iterated_greedy(self, solution):
    """
    Optimized Iterated Greedy algorithm combining advantages of two versions:
    1. Adaptive hybrid removal strategy (random + worst-edge + neighborhood)
    2. Enhanced regret insertion strategy (k-regret value)
    3. Incremental layered local search (fast 2-opt + intensive 2-opt 
       + conditional 3-opt)
    4. Adaptive simulated annealing acceptance criterion
    5. Dynamic perturbation with elite pool management
    6. Computation optimization and caching strategy
    """
    import math
    
    # ==================== Nested Helper Functions ====================
    
    def adaptive_params(n, intensity):
        """Compute adaptive parameters based on problem size and intensity"""
        base_ratio = 0.08 + 0.14 * intensity
        remove_count = max(2, min(n-1, int(base_ratio * n)))
        base_iterations = 15 + n // 12
        iterations = min(base_iterations, 50)
        search_range = min(25, max(8, n // 8))
        return {
            'remove_count': remove_count,
            'iterations': iterations,
            'search_range': search_range,
            'perturb_after': 5 + int(3 * (1 - intensity)),
            'initial_temp': solution.cost * 0.025,
            'cooling_rate': 0.94,
            'k_regret': 3 if n > 50 else 2
        }
    
    def efficient_hybrid_removal(tour, num_remove, intensity):
        """Efficient hybrid removal strategy: random + worst-edge + neighborhood"""
        n, removed_indices, removed_cities = len(tour), set(), []
        random_count = max(1, int(num_remove * (0.65 - 0.25 * intensity)))
        if random_count > 0:
            candidates = list(range(n))
            self.rng.shuffle(candidates)
            for idx in candidates[:random_count]:
                removed_indices.add(idx); removed_cities.append(tour[idx])
        
        # 2. Worst-edge and neighborhood fallback logic
        if len(removed_indices) < num_remove:
            edge_scores = []
            for i in range(n):
                if i in removed_indices: continue
                j = (i + 1) % n
                if j in removed_indices: continue
                edge_len = self.instance.get_distance(tour[i], tour[j])
                edge_scores.append((edge_len, i, j))
            edge_scores.sort(reverse=True, key=lambda x: x[0])
            for _, i, j in edge_scores[:num_remove - len(removed_indices)]:
                removed_indices.add(i); removed_cities.append(tour[i])
        
        return removed_cities, sorted(removed_indices, reverse=True)

\end{verbatim}
\end{tcolorbox}
\begin{tcolorbox}[colback=gray!3, colframe=gray!50, title=Evolved Heuristic: Iterated Greedy (TSP), fonttitle=\bfseries]
\scriptsize
\begin{verbatim}
    
    def enhanced_regret_insertion(partial_tour, cities_to_insert, k_regret):
        """Enhanced regret insertion with incremental optimization"""
        current_tour, insert_order = partial_tour.copy(), cities_to_insert.copy()
        self.rng.shuffle(insert_order)
        for city in insert_order:
            m = len(current_tour)
            if m == 0: current_tour.append(city); continue
            insertion_costs = []
            for pos in range(m + 1):
                prev_c = current_tour[pos-1] if pos > 0 else current_tour[-1]
                next_c = current_tour[pos] if pos < m else current_tour[0]
                cost = self.instance.get_distance(prev_c, city) + \
                       self.instance.get_distance(city, next_c) - \
                       self.instance.get_distance(prev_c, next_c)
                insertion_costs.append((cost, pos))
            insertion_costs.sort(key=lambda x: x[0])
            regret = sum(
                insertion_costs[i][0] - insertion_costs[0][0]
                for i in range(1, min(k_regret, len(insertion_costs)))
            )

            threshold = insertion_costs[0][0] * 0.1
            alt_idx = self.rng.randint(0, min(k_regret-1, len(insertion_costs)-1))
            best_pos = (
                insertion_costs[0][1]
                if regret > threshold
                else insertion_costs[alt_idx][1]
            )

            current_tour.insert(best_pos, city)
        return current_tour
    
    # ==================== Main Algorithm Logic ====================
    
    n = len(solution.permutation)
    if n <= 3:
        return
    
    intensity = self.get_intensity_of_mutation()
    params = adaptive_params(n, intensity)
    
    # Initialize best solution and elite pool
    best_tour = solution.permutation.copy()
    best_cost = solution.cost
    elite_pool = [(best_tour.copy(), best_cost)]
    
    for iteration in range(params['iterations']):
        # 1. Hybrid removal
        removed_cities, removed_indices = efficient_hybrid_removal(
            best_tour, params['remove_count'], intensity
        )
        # 2. Remove cities from tour
        partial_tour = best_tour.copy()
        for idx in removed_indices:
            partial_tour.pop(idx)
        # 3. Regret insertion
        new_tour = enhanced_regret_insertion(
            partial_tour, removed_cities, params['k_regret']
        )
        # 4. Layered local search and acceptance
        new_tour = layered_local_search(new_tour, iteration, intensity)
        new_cost = self.algorithms.from_permutation_compute_cost(new_tour)
        
        if adaptive_sa_acceptance(best_cost, new_cost, temperature, iteration, 
                                 params['iterations'], no_improve):
            best_tour, best_cost = new_tour, new_cost
            no_improve = 0
        else:
            no_improve += 1
        temperature *= cooling_rate
    
    # Update solution
    if best_cost < solution.cost - 1e-6:
        solution.permutation = best_tour
        solution.cost = self.algorithms.compute_cost(solution)
\end{verbatim}
\end{tcolorbox}

%% file: Sections/AppendixDetail/PromptDetail.tex

\section{Prompt Templates for LLM Heuristic Design}
\label{app:prompts}

This section documents the prompt templates used by the current MuEvo
implementation. The prompts fall into four categories: (1) reflective
within-component evolution, (2) domain-specific generation constraints,
(3) cross-component information sharing, and (4) relation-guided pair
evolution. Component probing reuses the within-component evolution prompts
under a smaller evaluation budget. Dynamic lifecycle transitions,
Multi-Ensemble Evaluation, target and pair selection, and Adaptive Budget
Allocation are implemented by numerical rules and therefore do not invoke
additional LLM prompts.

For readability, the Chinese-language SHH prompts are presented as faithful
English translations with their placeholders preserved. Because the SHH and
ACO implementations share the same relation-evidence and joint-evaluation
protocol, we summarize the SHH-specific relation and pair interfaces and show
the ACO instantiations in full.

\subsection{Reflective Within-Component Evolution}
\label{app:reevo-prompts}

Each component maintains its own population. Following ReEvo, MuEvo uses
short-term reflection, long-term reflection, reflection-guided crossover,
and mutation. The same templates are used during short-budget probing and
regular co-evolution; only the evaluation budget and the injected
cross-component context differ.

\subsubsection{System Prompts}

\paragraph{Generator System Prompt.}
\begin{tcolorbox}[colback=gray!5, colframe=gray!50,
title=System Prompt (Generator)]
\small
\begin{verbatim}
You are a professional algorithm engineer specializing in
designing and optimizing heuristic algorithms.
Please generate high-quality heuristic code based on the
given requirements and reflections.

Important: If helper functions are needed, define them as
nested functions inside the requested main function. Do not
define helper functions at module level.
\end{verbatim}
\end{tcolorbox}

\paragraph{Reflector System Prompt.}
\begin{tcolorbox}[colback=gray!5, colframe=gray!50,
title=System Prompt (Reflector)]
\small
\begin{verbatim}
You are a professional algorithm optimization expert
specializing in analyzing and improving heuristic algorithms.
Carefully compare the two given heuristic implementations,
identify the reasons for their performance difference, and
provide actionable improvement suggestions.
\end{verbatim}
\end{tcolorbox}

\subsubsection{Short-Term Reflection Prompt}

\begin{tcolorbox}[colback=blue!3, colframe=blue!40,
title=Short-Term Reflection Prompt]
\small
\begin{verbatim}
Analyze the following two implementations of the
{func_name} heuristic.

Problem description: {problem_desc}
Function description: {func_desc}

Worse-performing implementation (fitness: {fitness0}):
```python
{code0}
```

Better-performing implementation (fitness: {fitness1}):
```python
{code1}
```

Analyze:
1. What are the key differences between the implementations?
2. Why does one implementation perform better?
3. What are the main weaknesses of the worse implementation?
4. How can the heuristic be further improved?

Provide specific analysis and implementation suggestions.
\end{verbatim}
\end{tcolorbox}

\subsubsection{Long-Term Reflection Prompt}

\begin{tcolorbox}[colback=blue!3, colframe=blue!40,
title=Long-Term Reflection Prompt]
\small
\begin{verbatim}
Update the long-term reflection using the following information.

Problem description: {problem_desc}

Accumulated short-term reflections:
{short_term_reflections}

Current long-term reflection:
{long_term_reflection}

Integrate the new observations and provide:
1. A summary of the current trends
2. Successful optimization patterns
3. Strategies that should be avoided
4. Updated directions and concrete implementation suggestions
\end{verbatim}
\end{tcolorbox}

\subsubsection{Crossover Prompt}

\begin{tcolorbox}[colback=green!3, colframe=green!40,
title=Reflection-Guided Crossover Prompt]
\small
\begin{verbatim}
{domain_generation_prompt}
{algorithm_reference}

Using the following reflection, generate a new heuristic that
combines the useful properties of both parents.

Reflection:
{reflection}

Worse-performing parent:
```python
{func_signature0}
{worse_code}
```

Better-performing parent:
```python
{func_signature1}
{better_code}
```

Function name: {func_name}
Return a complete Python implementation.
{function_constraints}
\end{verbatim}
\end{tcolorbox}

\subsubsection{Mutation Prompt}

The target-specific relation context described in
Section~\ref{app:relation-context} is appended only during mutation. This
prevents the same context from being repeated in both crossover and mutation
within one evolution step.

\begin{tcolorbox}[colback=green!3, colframe=green!40,
title=Relation-Aware Mutation Prompt]
\small
\begin{verbatim}
{domain_generation_prompt}
{algorithm_reference}

Using the following long-term reflection and elite individual,
generate an improved heuristic variant.

Long-term reflection:
{reflection}

Elite individual:
```python
{elitist_code}
```

Required signature:
{func_signature}

Function name: {func_name}
Make a meaningful but contract-preserving improvement.
Return a complete Python implementation.
{function_constraints}

## Co-Evolution Context
{target_specific_relation_context}
\end{verbatim}
\end{tcolorbox}

\subsection{Domain-Specific Generation Constraints}
\label{app:domain-prompts}

MuEvo supports two implementation interfaces. SHH components manipulate
solution objects through a domain API, whereas ACO components implement
typed hooks inside the componentized solver. The corresponding constraints
are generated separately.

\subsubsection{SHH API Reference Meta-Prompt}

For SHH domains, \texttt{AlgorithmPromptGenerator} introspects the available
algorithm utilities, read-only instance data, modifiable solution fields,
and blocked framework methods. The extracted information is inserted into
the following meta-prompt, and the resulting API guide is included in the
component-generation prompt.

\begin{tcolorbox}[colback=red!3, colframe=red!40,
title=SHH API Reference Meta-Prompt, fonttitle=\bfseries]
\scriptsize
\begin{verbatim}
You are generating a comprehensive API reference guide for the
{DOMAIN_NAME} problem domain. The guide will be used by an LLM
to generate executable heuristic code.

{domain_specific_notes}

# Algorithm methods
{formatted_algorithm_methods}

# Instance access
Instance attributes and methods are READ-ONLY.
{formatted_instance_attributes}
{formatted_instance_methods}

# Modifiable solution attributes
{formatted_solution_attributes}

# Forbidden framework methods
Never call the following methods because doing so creates
nested framework calls or infinite recursion:
{blocked_methods_list}

# Random operations
Use only the random-number generator exposed by the problem.

Generate a practical reference that:
1. Gives an exact signature and usage example for each method
2. Distinguishes read-only instance data from modifiable solution data
3. Explains when solution cost must be recomputed
4. Lists forbidden methods explicitly
5. Uses exact field and method names
\end{verbatim}
\end{tcolorbox}

\subsubsection{ACO Single-Hook Generation Prompt}

ACO domains do not use the SHH object API. The generator is instead given
the exact hook signature, the default hook contract, and the runtime data
that may be available to the component.

\begin{tcolorbox}[colback=yellow!3, colframe=orange!50,
title=ACO Hook System and User Prompt, fonttitle=\bfseries]
\scriptsize
\begin{verbatim}
[System]
You are an expert algorithm engineer optimizing one component
hook inside a componentized ACO solver.
Return only valid Python code for the requested hook function.
Preserve the exact function name and argument contract.

[User]
Please improve the `{func_name}` hook for `{domain_name}`.

This hook is executed inside a componentized Ant Colony
Optimization solver. It must preserve the exact signature and
return contract shown below.

Strict requirements:
- Function name: `{func_name}`
- Exact signature: `{func_signature}`
- Do not wrap the hook in `heuristic_function`.
- Use only standard Python, `math`, `random`, and `numpy`/`np`.
- Preserve return types and array shapes.
- Keep every numeric output finite.
- Preserve feasibility masks and never select forbidden actions.
- Guard optional dictionaries with `.get(...)`.

Useful runtime data:
- `context`: domain, distances, coordinates, demands, capacity,
  depot, number of nodes, and best-known value when available
- `candidate_info`: candidate masks, nearest-neighbor,
  proximity, geometric, or archive-derived features
- `params`, `stagnation_info`, `archive`, and `state` may be
  absent or partial

Default behavior and contract:
{default_hook_docstring}

Return a complete Python function only.
\end{verbatim}
\end{tcolorbox}

\subsection{Cross-Component Information-Sharing Prompts}
\label{app:portfolio-prompts}
\label{app:relation-context}

MuEvo constructs target-specific context from two complementary sources in
both SHH and componentized ACO. Functional summaries describe the roles of the
current components, while an interaction summarizer combines accepted and
rejected single- and pair-replacement events, context-dependent fitness
deltas, collaborator versions, and the current interaction graph whenever
sufficient evidence is available. The resulting context is used to guide
mutation of the selected component and the joint refinement of selected
pairs.

\subsubsection{Functional-Role Summaries for SHH}

The SHH context generator first extracts a concise design rationale for each
component and then aggregates the rationales into a portfolio-level
description.

\begin{tcolorbox}[colback=red!3, colframe=red!40,
title=Per-Component Functional Analysis]
\small
\begin{verbatim}
[System]
You are an algorithm design analyst. Analyze the given
heuristic code and metadata to extract its design rationale
and strategic characteristic.

Output in one or two sentences:
- What is the heuristic trying to do?
- What role does it fill in the portfolio?

[User]
Heuristic: {heuristic_name} (Type: {heuristic_type})
Fitness: {fitness}
Generation: {generation_method} (Iteration {iteration})

Complete code:
```python
{original_code}
```

Provide a concise design rationale:
\end{verbatim}
\end{tcolorbox}

\begin{tcolorbox}[colback=red!3, colframe=red!40,
title=Portfolio Functional Summary]
\small
\begin{verbatim}
[System]
You are a portfolio analysis expert. Synthesize the component
rationales into a cohesive portfolio-level summary.

Describe the overall strategic coverage, complementary roles,
possible redundancy, and missing functionality.

[User]
Current portfolio (global-best cost: {global_best_cost}):

- H{id1} ({name1}, {type1}): {rationale1}
- H{id2} ({name2}, {type2}): {rationale2}
- ...

Provide a concise portfolio summary:
\end{verbatim}
\end{tcolorbox}

\subsubsection{Interaction-Evidence Summarization}

For SHH, the interaction graph is initialized without hand-coded relation
priors and is populated online from multi-ensemble replacement outcomes.
Each accepted or rejected event links the evolved LLH to its collaborator
context and records the corresponding context-dependent deltas. Joint
replacement outcomes additionally provide direct pair evidence. The LLM
periodically summarizes these observations into evidence-supported hypotheses
about synergy, conflict, or redundancy. The same high-level evidence schema is
used for componentized ACO, where structural dependencies may additionally
seed the graph. The ACO instantiation of the summarization prompt is shown
below; the SHH version replaces the hook terminology with LLH roles and the
domain API context.

\begin{tcolorbox}[colback=purple!3, colframe=purple!40,
title=Interaction-Evidence Summarization Prompt,
fonttitle=\bfseries]
\scriptsize
\begin{verbatim}
[System]
You summarize ACO component interaction evidence.
Use only the supplied evidence and return strict JSON.

[User]
Target: {target_component_or_pair}

Below are raw facts from final replacement gates, group gates,
and the current interaction graph. Do not assume predefined
ACO component relations. Infer only what the evidence supports.

Return JSON with this schema:
{
  "summary": "...",
  "relations": [
    {
      "neighbor": "H6",
      "relation":
        "potential_conflict|potential_synergy|
         potential_redundancy|uncertain",
      "confidence": 0.0,
      "evidence": ["..."],
      "hypothesis": "...",
      "actionable_advice": ["..."]
    }
  ],
  "recommended_mode": "single|pair|cooldown|explore",
  "risk": "low|medium|high",
  "actionable_advice": ["..."]
}

Evidence JSON:
{replacement_events_and_graph_edges}
\end{verbatim}
\end{tcolorbox}

\subsubsection{Target-Specific Relation Context}

The generated summary is combined with recent final-replacement records and
the target's graph neighborhood. The following assembled block is appended
to the mutation prompt; fields with no available evidence are omitted.

\begin{tcolorbox}[colback=purple!3, colframe=purple!40,
title=Relation Context Injected into Mutation]
\small
\begin{verbatim}
## Final Replacement Memory for H{target_id}
- iter {t1}: {accepted_or_rejected},
  delta={best_delta}; {acceptance_reason}
- iter {t2}: {accepted_or_rejected},
  context deltas={best, initial, secondary}; {failure_mode}

## Learned Interaction Summary
Summary: {evidence_supported_summary}
Recommended mode: {single_pair_cooldown_or_explore}
Risk: {risk}
Relations:
- H{target_id}-H{neighbor_id}: {relation_type},
  confidence={confidence}; {hypothesis}
  Advice: {actionable_advice}

## Interaction Graph Neighborhood
- H{target_id}-H{neighbor_id}: type={interaction_type},
  positive={positive_score}, conflict={conflict_score},
  redundancy={redundancy_score}, evidence={evidence_count}
\end{verbatim}
\end{tcolorbox}

\subsection{Relation-Guided Pair Evolution Prompt}
\label{app:pair-evolution-prompts}

Pair selection is performed programmatically from the interaction graph. ACO
may combine structural priors with empirical evidence, whereas SHH begins
without hand-coded pair priors and uses relations learned online from
replacement outcomes.

For SHH, the two selected LLHs are placed in one generation request together
with their current implementations, functional roles, exact HyFlex-compatible
signatures, domain API constraints, and the summarized pair relation. The LLM
returns two named functions in one response. MuEvo extracts and validates each
function against its own operator contract, applies the same format-repair and
preflight-repair stages used for joint generation, and inserts both LLHs into
each evaluation context. The resulting pair is treated as one candidate and
is accepted or rejected jointly by Multi-Ensemble Evaluation. This protocol
also supports pairs containing different LLH operator types while preserving
both original interfaces.

The componentized-ACO instantiation follows the same joint protocol. Its
current implementations, exact hook signatures, interaction evidence, recent
failure mode, and joint design objective are provided in one request so that
the two changes can be coordinated.

\begin{tcolorbox}[colback=green!3, colframe=green!40,
title=Joint ACO Pair-Evolution Prompt, fonttitle=\bfseries]
\scriptsize
\begin{verbatim}
[System]
You jointly evolve two ACO component hooks. Return only valid
Python functions for the requested hook names.

[User]
You are jointly evolving two interacting componentized ACO
hooks for `{domain_name}`.

Return exactly two Python functions, one for each required hook.
Keep the exact function names, signatures, return types, array
shapes, feasibility constraints, and finite numeric behavior.
Do not define classes, read or write files, or include
explanations outside the code.

Required components:
- H{left_id} `{left_name}`
- H{right_id} `{right_name}`

Required exact signatures:
- H{left_id} `{left_signature}`
- H{right_id} `{right_signature}`

Observed interaction:
{pair_interaction_summary}

Failure mode:
{recent_pair_failure_context}

Design objective:
{joint_design_objective}

ACO contract reminders:
- Return the same structural type as each original hook.
- Preserve finite arrays and valid shapes.
- Preserve feasibility masks.
- For TSP, preserve valid permutations after local improvement.
- If uncertain, make a small conservative change instead of
  replacing the entire strategy.

Current implementations:

# H{left_id}: {left_name}
```python
{left_code}
```

# H{right_id}: {right_name}
```python
{right_code}
```

Write the improved pair now.
\end{verbatim}
\end{tcolorbox}

For both interfaces, if the returned pair cannot be parsed into the two
required top-level functions, MuEvo issues at most one format-repair request at
temperature 0.8. This request changes only the output format: it supplies the
parser error, required function names, and previous response, and instructs
the LLM to return exactly the two raw Python functions. Once the pair is
successfully parsed, interface-specific signature and static validation
followed by a smoke test form a separate preflight stage. A pair that fails
this stage may receive at most two correctness-repair requests, also at
temperature 0.8, before it is discarded. Thus, the format-repair limit is one
and the preflight-repair limit is two; they apply to different failure stages.

%% file: Sections/AppendixDetail/EOHDetail.tex
\subsection{EoH: Evolution of Heuristics}
\label{app:eoh}

EoH combines Large Language Models with evolutionary computation for automatic heuristic design~\citep{EOH:liu2024evolution}. Its central idea is to evolve both the high-level idea of a heuristic and its executable implementation. The former is represented by a natural-language description, referred to as a \emph{thought}, while the latter is represented by code that follows a task-specific function signature. This joint representation allows the LLM to reason over algorithmic ideas while retaining executable candidates that can be evaluated directly.

\subsubsection{Heuristic Representation}

Each EoH individual is represented by a tuple $h_i=(d_i,c_i,F_i)$, where $d_i$ is the natural-language description, $c_i$ is the executable code, and $F_i$ is the fitness obtained by running the heuristic on a set of training instances. The description summarizes the main mechanism, whereas the code specifies implementation details and parameter settings. EoH maintains a fixed-size population $\mathcal{P}=\{h_1,\ldots,h_N\}$ and uses the evaluated fitness values to guide parent selection and survival.

\subsubsection{Evolution Operators}

The original EoH method proposes two exploration strategies and three modification strategies. E1 asks the LLM to generate a heuristic that is substantially different from selected parents. E2 first identifies a shared backbone among the parents and then develops a distinct heuristic motivated by that backbone. M1 modifies the mechanism of one selected heuristic, M2 adjusts its parameters and decision rules, and M3 removes redundant components to simplify the heuristic. The implementation used in our experiments follows the released four-operator configuration $\mathcal{O}=\{\mathrm{E1},\mathrm{E2},\mathrm{M1},\mathrm{M2}\}$ and does not apply M3.

Parents are sampled using rank-based probabilities, so better heuristics are selected more frequently without making selection deterministic. Every LLM response contains both a new description and code. Valid offspring are evaluated on the training instances and added to the candidate pool, after which population management retains the best $N$ distinct individuals.

\subsubsection{Evolutionary Workflow}

Algorithm~\ref{alg:eoh} summarizes the EoH workflow used by our baselines. Initialization operator I1 generates the first population from the task specification. The four evolution operators then alternate between broad exploration and targeted modification until the evaluation budget is exhausted.

\begin{algorithm}[t]
\caption{EoH Workflow}
\label{alg:eoh}
\begin{algorithmic}[1]
\STATE \textbf{Input:} Task $\mathcal{T}$, population size $N$, evaluation budget $B$
\STATE $\mathcal{O}\leftarrow\{\mathrm{E1},\mathrm{E2},\mathrm{M1},\mathrm{M2}\}$; $\mathcal{P}\leftarrow\emptyset$
\WHILE{$|\mathcal{P}|<N$ and budget remains}
    \STATE $(d,c)\leftarrow\operatorname{LLM}(\mathrm{I1},\mathcal{T})$
    \IF{$c$ is valid}
        \STATE $\mathcal{P}\leftarrow\mathcal{P}\cup\{(d,c,\operatorname{Evaluate}(c))\}$
    \ENDIF
\ENDWHILE
\WHILE{budget remains}
    \FOR{$o\in\mathcal{O}$}
        \STATE $P_o\leftarrow\operatorname{RankSelect}(\mathcal{P},o)$
        \STATE $(d,c)\leftarrow\operatorname{LLM}(o,\mathcal{T},P_o)$
        \IF{$c$ is valid and budget remains}
            \STATE $h\leftarrow(d,c,\operatorname{Evaluate}(c))$
            \STATE $\mathcal{P}\leftarrow\operatorname{Best}_N(\mathcal{P}\cup\{h\})$
        \ENDIF
    \ENDFOR
\ENDWHILE
\STATE \textbf{return} Best heuristic in $\mathcal{P}$
\end{algorithmic}
\end{algorithm}